\documentclass[10pt]{book}

\usepackage[table]{xcolor}
\usepackage{mathrsfs} 
\usepackage{amssymb,amsmath}
\usepackage{graphicx}
\usepackage{hyperref} 
\usepackage{xspace}
\usepackage{booktabs}
\usepackage{caption} 
\usepackage{longtable}

\usepackage{pdfpages}
\usepackage{caption}

\usepackage[T1]{fontenc}

\hypersetup{colorlinks,linkcolor={red!50!black},citecolor={blue!50!black},urlcolor={blue!80!black}}
\usepackage[top=1.2in, bottom=1.2in, left=1.4in, right=1.4in]{geometry}

\usepackage{tikz}
\usetikzlibrary{shapes,calc,positioning,automata,arrows,trees}

\newcommand\bcpen{\includegraphics[width=15pt]{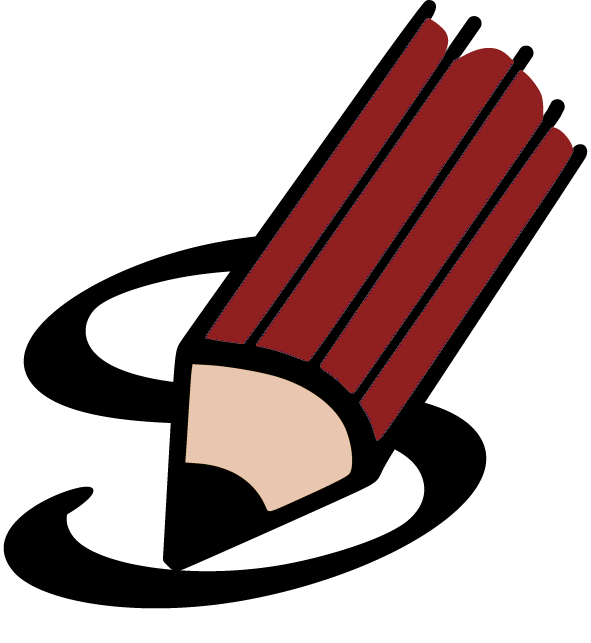}}

\usepackage{listingsutf8}
\usepackage{textcomp}
\usepackage{verbatim}
\usepackage{multirow}

\definecolor{v3lgray}{gray}{0.98}
\definecolor{v2lgray}{gray}{0.85}
\definecolor{vlgray}{gray}{0.92}
\definecolor{dgray}{rgb}{0.4,0.4,0.4}
\definecolor{dblue}{RGB}{0,0,99}
\definecolor{dred}{RGB}{175,6,54}
\definecolor{lred}{RGB}{155,6,44}
\definecolor{dgreen}{RGB}{47,135,7}
\definecolor{dviolet}{RGB}{102,0,153}
\definecolor{mblue}{RGB}{0,0,180}
\definecolor{dorange}{RGB}{204, 82, 0}

\def\ace{ACE\xspace}

\def\x3{{\rm XCSP$^3$}\xspace}
\def\jv3{{\rm JvCSP$^3$}\xspace}
\def\p3{{\rm PyCSP$^3$}\xspace}

\newcommand{\gb}[1]{{\tt #1}} 

\newtheorem{remark}{Remark}

\lstdefinelanguage{json}{
    basewidth  = {.6em,0.6em},
    basicstyle=\normalfont\ttfamily,
    breaklines=true,
    morestring=[b]',
    morestring=[b]", 
    sensitive=false,
    stringstyle=\color[rgb]{0.227,0.226,0.441}\ttfamily, 
    escapechar=!,
    showstringspaces=false,
    xleftmargin=20pt, 
    breaklines=true,basicstyle=\ttfamily\small,inputencoding=utf8/latin9,texcl
}
\lstnewenvironment{json}{\lstset{language=json}}{}

\lstdefinelanguage{mcsp}{
  keywords={forall,array,block,class,implements,model,public,slide},
  basewidth  = {.6em,0.6em},
  keywordstyle=\color{dred}\bfseries,
  ndkeywords={intension,lessThan,lessEqual,greaterEqual,greaterThan,equal,different,implication,equivalence,conjunction,disjunction,extension,regular,mdd,allDifferent,allDifferentMatrix,allEqual,ordered,increasing,decreasing,strictlyIncreasing,strictlyDecreasing,lex,lexMatrix,sum,count,atMost,atLeast,exactly,atMost1,atleast1,exactly1,element,channel,maximum,minimum,cardinality,nValues,noOverlap,cumulative,instantiation,clause,circuit,minimize,maximize},
  ndkeywordstyle=\color{mblue}\bfseries,
  identifierstyle=\color{black},
  sensitive=false,
  comment=[l]{//},
  morecomment=[s]{<!--}{-->},
  commentstyle=\color{dgreen}\ttfamily,
  stringstyle=\color{dgreen}\rmfamily, 
  morestring=[b]',
  morestring=[b]",
  escapechar=~,
  showstringspaces=false,
  classoffset=2, morekeywords={private},keywordstyle=\color{gray},
  classoffset=3, morekeywords={dom,size,when},keywordstyle=\color{dorange},
  xleftmargin=-22pt,xrightmargin=-22pt,
  breaklines=true,basicstyle=\ttfamily\footnotesize,backgroundcolor=\color{v3lgray},inputencoding=utf8/latin9,texcl
}
\lstnewenvironment{mcsp}{\lstset{language=mcsp}}{}

\definecolor{colorex}{RGB}{255,248,220} 
\definecolor{mgray}{rgb}{0.55,0.55,0.55}
\definecolor{officegreen}{rgb}{0.0, 0.5, 0.0}

\newcounter{cntPy}

\lstnewenvironment{python}{\lstset{
    inputencoding=utf8/latin1,
    language=python,
    ndkeywords={size,dom,matrix,strict,occurrences},
    ndkeywordstyle=\color{mblue}, 
    keywords=[2]{data,satisfy,minimize,maximize,solve}, 
    keywordstyle=[2]\color{dred}, 
    deletekeywords={def},
    keywords=[3]{def,If,Then,Else,Match}, 
    keywordstyle=[3]\color{dorange},
    commentstyle=\color{officegreen}, 
    showstringspaces=false,
    captionpos=t,
    basicstyle=\ttfamily\footnotesize,
    framesep=5pt,
    xleftmargin=5pt,xrightmargin=0pt,
    backgroundcolor=\color{colorex}, 
    escapechar=@
  }
}{}

\usepackage[skins,breakable,xparse]{tcolorbox}
\newcommand{\core}[1]{ 
  \medskip \begin{tcolorbox}[
    enhanced,breakable,
    boxsep=0pt,top=4pt,bottom=0pt,left=2mm,right=1mm,
    toprule=0.1mm,leftrule=0.1mm,rightrule=0.25mm,bottomrule=0.25mm,shadow={0.2mm}{-0.2mm}{0mm}{dgray},
    overlay unbroken and first={\node (logo) at ([xshift=4mm,yshift=-5mm]frame.north west) {#1}; },
    colframe=dgray,titlerule=-0.2mm,toptitle=3mm,coltitle=dred, fonttitle=\bfseries,
    lines before break=6, pad at break*=10pt
    }

\newenvironment{boxpy}
 {\stepcounter{cntPy} \core{\bcpen} , colback=colorex, title style={color=colorex}, title=~ ~ \,PyCSP$^3$ Model \thecntPy]}
 {\end{tcolorbox}}

\newenvironment{command}
  {\quote\small\verbatim}
  {\endverbatim\endquote}

  \usepackage{authblk}
 
\title{\textcolor{dred}{Proceedings of the \x3 Competition 2025}}
\author{Gilles Audemard \and Christophe Lecoutre \and Emmanuel Lonca}

\affil{\vspace{1cm} CRIL \\University of Artois \& CNRS \\ France}

\date{November 3, 2025}

\begin{document}
\maketitle

~ \\
~ \\

\bigskip

This document represents the proceedings of the \x3 Competition 2025, following those published in 2022 \cite{compet22}, 2023 \cite{compet23}, and 2024 \cite{compet24}.
The website containing all {\bf detailed results} of this international competition is available at:
\begin{quote}
  \href{https://www.cril.univ-artois.fr/XCSP25/}{https://www.cril.univ-artois.fr/XCSP25}
\end{quote}

  \bigskip
  \bigskip
\noindent The organization of this 2025 competition involved the following tasks:
\begin{itemize}
\item adjusting general details (dates, tracks, $\dots$) by G. Audemard, C. Lecoutre and E. Lonca
\item selecting instances (problems, models and data) by C. Lecoutre
\item receiving, testing and executing solvers on CRIL cluster by E. Lonca
\item validating solvers and rankings by C. Lecoutre and  E. Lonca
\item developping the 2025 website dedicated to results by G. Audemard
\end{itemize}

\bigskip\bigskip 
{\bf Important}: for reproducing the experiments and results, it is important to use the very same set of \x3 instances, as used in the competition.
These instances can be found in this \href{https://www.cril.univ-artois.fr/~lecoutre/compets/instancesXCSP25.zip}{archive}.
Some (usually minor) differences may exist when compiling the models presented in this document and those that can be found in this  \href{https://www.cril.univ-artois.fr/~lecoutre/compets/modelsXCSP25.zip}{archive}.


\tableofcontents

\chapter{About the Selection of Problems in 2025}

Remember that the complete description, {\bf Version 3.2}, of the format (\x3) used to represent combinatorial constrained problems can be found in \cite{xcsp3}.
As usual for \x3 competitions, we have limited \x3 to its kernel, called \x3-core \cite{xcsp3core}.
This means that the scope of the \x3 competition is restricted to:
\begin{itemize}
\item integer variables
\item instances of  frameworks:
  \begin{itemize}
  \item CSP (Constraint Satisfaction Problem)
  \item COP (Constraint Optimization Problem)
    \end{itemize}
\item a set of 24 popular (global) constraints for main tracks:
  \begin{itemize}
  \item generic constraints: \gb{intension} and \gb{extension} (also called \gb{table})
  \item language-based constraints: \gb{regular} and \gb{mdd}
  \item comparison constraints: \gb{allDifferent}, \gb{allDifferentList}, \gb{allEqual}, \gb{ordered}, \gb{lex} and \gb{precedence}
  \item counting/summing constraints: \gb{sum}, \gb{count}, \gb{nValues} and \gb{cardinality}
  \item connection constraints: \gb{maximum}, \gb{minimum}, \gb{element} and \gb{channel}
  \item packing/scheduling constraints: \gb{noOverlap}, \gb{cumulative}, \gb{binPacking} and \gb{knapsack}
  \item \gb{circuit}, \gb{instantiation} (and \gb{slide})
  \end{itemize}
  and a smaller set of constraints for mini tracks.
\end{itemize}

\medskip
For the 2025 competition, 33 problems have been selected.
They are succinctly presented in Table \ref{fig:problems}.
For each problem, the type of the involved (global) constraints is indicated.
At this point, do note that making a good selection of problems/instances is a difficult task.
In our opinion, important criteria for a good selection are:
\begin{itemize}
\item the novelty of problems, avoiding constraint solvers to overfit already published problems;
\item the diversity of constraints, trying to represent all of the most popular constraints (those from \x3-core) while paying attention to not over-representing some of them;
\item the scaling up of problems.
\end{itemize}

\begin{table}
  \begin{tabular}{p{5.5cm}p{0.2cm}p{7.5cm}}
    \toprule
CSP Problems & & Global Constraints)\\
    \midrule
\rowcolor{vlgray}{}  Accordion & & \gb{element}  \\
    AlmostMagic &   &   \gb{allDifferent}, \gb{sum} \\
\rowcolor{vlgray}{}  ChainReaction   &  & \gb{allDifferent}  \\
    CrazyFrog & &  \gb{allDifferent}, \gb{circuit}  \\
\rowcolor{vlgray}{}   Cryptanalysis & &  \gb{sum} \\
    EFPA & & \gb{cardinality}, \gb{count}, \gb{lex} \\
\rowcolor{vlgray}{}  GracefulGraph & & \gb{allDifferent}\\
Heterosquare & &  \gb{allDifferent}, \gb{sum} \\
\rowcolor{vlgray}{} LangfordBin & & \gb{element} \\
LotteryDesign & & \gb{allDifferentList} \\
\rowcolor{vlgray}{} PegSolitaireTable & & \gb{table} \\
RamseyPartition &  & \gb{cardinality}, \gb{nValues} \\
\rowcolor{vlgray}{} Rostering & & \gb{allDifferent}, \gb{regular} \\
RotatingRostering &  & \gb{allEqual}, \gb{cardinality}, \gb{count}, \gb{table} \\
\rowcolor{vlgray}{} SEDF & & \gb{allDifferent}, \gb{cardinality}, \gb{element} \\
TilingRythmicCanons &  & \gb{allDifferent}, \gb{nValues} \\
& & \\
\midrule
COP Problems & & Global Constraints \\ 
\midrule
\rowcolor{vlgray}{} AlteredStates &   & \gb{element}, \gb{lex}, \gb{sum} \\
 BlockModeling &   &  \gb{sum} \\
\rowcolor{vlgray}{} BusScheduling &   & \gb{count}, \gb{sum} \\ 
ButtonsScissors &  & \gb{count}, \gb{maximum} \\
\rowcolor{vlgray}{} ChampionsLeague &   & \gb{element}, \gb{sum}, \gb{table} \\
Coprime &  & \gb{table} \\
\rowcolor{vlgray}{} CutStock &  & \gb{lex}, \gb{sum} \\
FAPP &  & \gb{table}  \\
\rowcolor{vlgray}{} FlexibleJobshopScen &  & \gb {count}, \gb{cumulative}, \gb{sum} \\
Fortress &   & \gb{sum}, \gb{table$^*$}  \\
\rowcolor{vlgray}{} IHTC &  & \gb{binPacking}, \gb{cumulative}, \gb{element}, \gb{sum}, \gb{table} \\
LowAutocorrelation &  & \gb{sum} \\
\rowcolor{vlgray}{} MetabolicNetwork &  & \gb{sum} \\
RoadefPlanning &  &  \gb{cardinality}, \gb{sum}, \gb{table} \\
\rowcolor{vlgray}{} RollerSplat &   & \gb{element}, \gb{table$^*$} \\
SchedulingOS &  & \gb{allDifferent}, \gb{element}, \gb{maximum}, \gb{noOverlap}  \\
\rowcolor{vlgray}{} TankAllocation1 & & \gb{binPacking}, \gb{sum}, \gb{table} \\
TankAllocation2 &  & \gb{allDifferent}, \gb{count}, \gb{sum}, \gb{table} \\
\bottomrule
  \end{tabular}
  \caption{Selected Problems for the main tracks of the 2025 Competition. When \gb{table} is followed by ($*$), it means that starred tables are involved.}\label{fig:problems}
\end{table}

\paragraph{Novelty.} Almost all problems are new in 2025, with models directly written in \p3.
Instances of problem ``Metabolic Network'' have been proposed by Maxime Mahout and François Fages.



\paragraph{Scaling up.} It is always interesting to see how constraint solvers behave when the instances of a problem become harder and harder.
This is what we call the scaling behavior of solvers.
For most of the problems in the 2025 competition, we have selected series of instances with regular increasing difficulty.
It is important to note that assessing the difficulty of instances was mainly determined with \ace \cite{ace}, which is the reason why \ace is declared to be off-competition (due to this strong bias).

\paragraph{Selection.} This year, the selection of problems and instances has been performed by Christophe Lecoutre.
As a consequence, the solver \ace was labeled off-competition.


\bigskip

\chapter{Problems and Models}

In the next sections, you will find all models used for generating the \x3 instances of the 2025 competition (for main CSP and COP tracks).
Almost all models are written in \p3 \cite{pycsp3}, Version 2.5 (released in June 11, 2025); see \href{https://pycsp.org}{https://pycsp.org}.

\section{CSP}

\subsection{Accordion}

\paragraph{Description.}
From \cite{tabID}:
\begin{quote}
  Accordion (BVS Development Corporation, 2017) is a single-player (patience or solitaire) card game.
    The game starts with the chosen cards in a sequence, each element of which we consider as a 'pile' of one card.
    Each move we make consists of moving a pile on top of either the pile immediately to the left, or three to the left
    (i.e. with two piles between the source and destination) such that the top cards in the source and destination piles match
    by either rank (value of the card, e.g. both 7) or suit (clubs, hearts, diamonds, or spades).
    The result of each move is to reduce the number of piles by 1 and change the top card of the destination pile.
    The empty space left at the position of the source pile is deleted.
    The goal is to keep making moves until just one pile remains.
\end{quote}

\paragraph{Data.}

As an illustration of data specifying an instance of this problem, we have:
\begin{json}
{
  "cards": [17, 11, 49, 3, 18, 0, 10, 8, 42, 30, 6]
}
\end{json}

\paragraph{Model.}
The \p3 model, in a file `Accordion.py', used for the competition is: 

\begin{boxpy}\begin{python}
@\imp@

cards = data or load_json_data("11-01.json")

n, nSteps = len(cards), len(cards)

T, N = range(nSteps - 1), range(n)  # T used for iterating from 0 to nSteps-1 (excluded)

# pf[t] is the pile which is moved from at time t
pf = VarArray(size=nSteps - 1, dom=range(n))

# pt[t] is the pile which is moved to at time t
pt = VarArray(size=nSteps - 1, dom=range(n))

# cf[t] is the top card of the pile which is moved from at time t
cf = VarArray(size=nSteps - 1, dom=range(52))

# ct[t] is the top card of the pile which is moved to at time t
ct = VarArray(size=nSteps - 1, dom=range(52))

# x[t][i] is the top card of the jth pile at time t
x = VarArray(size=[nSteps, n], dom=range(52))

satisfy(
   # setting initial state
   x[0] == cards,

   # making the move
   [
      (
         x[t][pf[t]] == x[t + 1][pt[t]],
         cf[t] == x[t][pf[t]],
         ct[t] == x[t][pt[t]]
      ) for t in T
   ],

   # ensuring each move is either 1 or three places
   [
      either(
         pt[t] == pf[t] - 1,
         pt[t] == pf[t] - 3
      ) for t in T
   ],

   # ensuring that the top cards of the two involved piles are of the same rank or suit
   [
      either(
         cf[t] % 13 == ct[t] % 13,
         cf[t] // 13 == ct[t] // 13
      ) for t in T
   ],

   # ensuring unmoved cards are copied from one timestep to the next one
   [
      If(
         i < pt[t],
         Then=x[t][i] == x[t + 1][i]
      ) for t in T for i in N
   ] + [
      If(
         i > pt[t], i < pf[t],
         Then=x[t][i] == x[t + 1][i]
      ) for t in T for i in N
   ] + [
      If(
         i >= pf[t],
         Then=x[t + 1][i] == x[t][i + 1]
      ) for t in T for i in N[:-1]
   ], 

   # unused slots are filled up with zero
   x[1:, -1] == 0,

   [pf[t] <= n - t + 1 for t in T]
)
\end{python}\end{boxpy}

This model involves five arrays of variables, many intensional constraints and some constraints \gb{Element}.
A series of 12 instances has been selected for the competition.
For generating an \x3 instance (file), you can execute for example:
\begin{command}
  python Accordion.py -data=11-01.json
\end{command}

\subsection{Almost Magic}

\paragraph{Description.}

From \href{https://www.janestreet.com/puzzles/almost-magic-index/}{JaneStreet}:
\begin{quote}
  An almost magic square is, well, almost a magic square.
  It differs from a magic square in that the 8 sums may differ from each other by at most 1.
  For this puzzle, place distinct positive integers into the empty grid above such that each of four bold-outlined 3-by-3 regions is an almost magic square.
  Your goal is to do so in a way that minimizes the overall sum of the integers you use.
\end{quote}

\paragraph{Data.}

Two integers are required to specify a specific instance: the order $n$ of the board and the number $p$ of possible values.
The values of $(n,p)$ used for generating the 2025 competition instances are:
\begin{quote}
  (3, 30), (3, 35), (3, 40), (3, 70), (4, 50), (4, 80), (5, 70), (6, 100), (7, 130), (8, 160), \\(9, 200), (10, 250)
\end{quote}

\paragraph{Model.}
The \p3 model, in a file `AlmostMagic.py', used for the competition is: 

\begin{boxpy}\begin{python}
@\imp@

n, p = data or (3, 100)

N = range(1, n)

# x[k][i][j] is the value at row i and column j in the kth region
x = VarArray(size=[4, n, n], dom=range(1, p))

# y[k] is the almost magic value of the kth region
y = VarArray(size=4, dom=range(p * 3))
     
satisfy(
   # ensuring different values
   AllDifferent(
      x[0], x[1][0][1:], x[1][1][1:], x[1][2][2],
      x[2][0][0], x[2][1][:-1], x[2][2][:-1], x[3]
   ),

   # ensuring almost magic regions
   [
      (
         Sum(t) >= y[k],
         Sum(t) <= y[k] + 1
      ) for k in range(4) for t in rows(x[k]) + columns(x[k]) + diagonals(x[k])
    ],

   # dealing with overlapping cells
   [
      [x[0][i][-1] == x[1][i - 1][0] for i in N],
      [x[0][-1][j - 1] == x[2][0][j] for j in N],
      [x[1][-1][j - 1] == x[3][0][j] for j in N],
      [x[2][i][-1] == x[3][i - 1][0] for i in N]
    ]
)
\end{python}\end{boxpy}

This model involves two arrays of variables, a constraint \gb{AllDifferent}, many intensional constraints and many constraints \gb{Sum}.
A series of 12 instances has been selected for the competition.
For generating an \x3 instance (file), you can execute for example:
\begin{command}
  python AlmostMagic.py -data=[3,100] 
\end{command}

\subsection{Chain Reaction}

\paragraph{Description.}

From \href{https://www.janestreet.com/puzzles/chain-reaction-index/}{JaneStreet}:
\begin{quote}
Write down a chain of integers between 1 and $p$ (e.g., 100), with no repetition, such that if $a$ and $b$ are consecutive numbers in the chain,
then $a$ evenly divides $b$ or $b$ evenly divides $a$.
Here is an example of such a chain, with length 12: \\
$~ ~ ~ $ 37, 74, 2, 8, 4, 16, 48, 6, 3, 9, 27, 81 \\
\end{quote}

\paragraph{Data.}

Two integers are required to specify a specific instance: the length $n$ of the chain and the number $p$ of possible values.
The values of $(n,p)$ used for generating the 2025 competition instances are:
\begin{quote}
  (20, 20), (20, 25), (30, 35), (30, 40), (40, 50), (40, 55), (50, 65), (50, 70), \\
  (60, 85), (60, 90), (70, 95), (70, 100), (80, 105), (80, 110)
\end{quote}

\paragraph{Model.}
The \p3 model, in a file `ChainReaction.py', used for the competition is: 

\begin{boxpy}\begin{python}
@\imp@

n, k = data or (20, 100)

# x[i] is the ith value of the chain
x = VarArray(size=n, dom=range(1, k + 1))

satisfy(
   # ensuring all values are different
   AllDifferent(x),

   # ensuring that two consecutive numbers v and w of the chain satisfy the property 
   [
      either(
         x[i] % x[i + 1] == 0,
         x[i + 1] % x[i] == 0
      ) for i in range(n - 1)
   ]
)
\end{python}\end{boxpy}

This model involves an array of variables, a constraint \gb{AllDifferent}, and many intensional constraints.
A series of 14 instances has been selected for the competition.
For generating an \x3 instance (file), you can execute for example:
\begin{command}
  python ChainReaction.py -data=[20,25] 
\end{command}

\subsection{Crazy Frog}

\paragraph{Description.}
From \href{https://github.com/lpcp-contest/lpcp-contest-2021/tree/main/problem-1}{LPCP Contest 2021}:
\begin{quote}
The problem is based on the Crazy Frog Puzzle.
You have the control of a little frog, capable of very long jumps.
The little frog just woke up in an $n \times n$land, with few obstacles and a lot of insects.
The frog can jump as long as you want, but only in the four cardinal directions (don't ask why) and you cannot land on any obstacle or already visited places.
\end{quote}

\paragraph{Data.}

As an illustration of data specifying an instance of this problem, we have:
\begin{json}
{
  "frog": [3, 14],
  "grid": [
    [1, 1, 0, 0, 0, 0, 0, 0, 0, 0, 1, 0, 0, 0, 1, 0, 0, 0, 1, 1],
    [0, 0, 1, 1, 1, 0, 0, 0, 0, 1, 0, 0, 0, 0, 0, 0, 0, 1, 0, 0],
    [0, 0, 1, 1, 0, 0, 0, 1, 0, 0, 1, 1, 0, 0, 0, 0, 0, 0, 1, 0],
    [0, 1, 1, 1, 1, 1, 1, 0, 0, 0, 0, 0, 0, 1, 0, 0, 0, 1, 0, 0],
    [0, 0, 1, 0, 0, 0, 0, 1, 0, 0, 1, 1, 1, 0, 0, 0, 0, 0, 0, 1],
    [0, 0, 0, 1, 0, 0, 1, 1, 1, 1, 0, 0, 0, 0, 0, 0, 0, 1, 0, 1],
    [0, 1, 1, 1, 0, 0, 0, 0, 0, 0, 0, 0, 0, 1, 0, 0, 0, 0, 1, 0],
    [0, 0, 0, 1, 0, 0, 0, 0, 1, 0, 0, 1, 0, 0, 0, 0, 0, 0, 0, 0],
    [1, 0, 0, 0, 1, 0, 0, 0, 0, 1, 0, 0, 0, 0, 1, 0, 1, 0, 0, 1],
    [0, 0, 1, 1, 0, 0, 0, 0, 1, 0, 0, 0, 0, 0, 0, 0, 0, 0, 0, 0],
    [0, 1, 1, 0, 1, 0, 0, 0, 0, 0, 0, 0, 0, 0, 1, 0, 1, 0, 0, 1],
    [0, 0, 0, 0, 0, 0, 1, 1, 0, 1, 0, 1, 1, 0, 1, 0, 0, 0, 0, 0],
    [0, 0, 1, 0, 0, 1, 0, 1, 0, 0, 0, 0, 1, 0, 0, 0, 0, 0, 1, 0],
    [0, 1, 0, 0, 0, 0, 0, 0, 0, 0, 0, 0, 0, 0, 1, 0, 0, 0, 0, 0],
    [1, 0, 0, 0, 1, 0, 0, 0, 0, 0, 0, 1, 1, 0, 1, 0, 0, 0, 0, 1],
    [0, 1, 1, 0, 0, 0, 1, 0, 0, 1, 0, 0, 0, 0, 0, 0, 0, 1, 1, 0],
    [0, 0, 1, 0, 0, 0, 0, 0, 0, 0, 1, 1, 0, 0, 1, 0, 1, 1, 0, 0],
    [0, 0, 0, 0, 1, 0, 0, 0, 0, 0, 0, 0, 0, 0, 0, 0, 0, 0, 0, 0],
    [0, 0, 0, 0, 0, 0, 1, 0, 0, 1, 0, 0, 0, 1, 0, 1, 1, 0, 0, 0],
    [0, 1, 0, 1, 0, 0, 0, 1, 0, 0, 0, 0, 1, 0, 0, 1, 0, 0, 1, 0]
  ]
}
\end{json}

\paragraph{Model.}

The two variants (a main one, and another one called ``table'') of the \p3 model, in a file `CrazyFrog.py', used for the \x3 competition are:

\begin{boxpy}\begin{python}
@\imp@

frog, grid = data or load_json_data("06.json")

n = len(grid)

zeros = [(i,j) for i in range(n) for j in range(n) if grid[i][j] == 0]  
nZeros = len(zeros)

comp_zeros = [ # compatible zeros
  c(k, q) for k in range(nZeros) for q in range(nZeros)
     if k != q and (zeros[k][0] == zeros[q][0] or zeros[k][1] == zeros[q][1])
]

if not variant():
   # x[i] is the zero that follows the ith zero
   x = VarArray(
       size=nZeros,
       dom=lambda i: {q for k, q in comp_zeros if k == i} | ({0} if i != 0 else set())
   )

   satisfy(
      # ensuring a circuit
      Circuit(x),
   )

elif variant('table'):

   # x[i] is the ith zero visited by the frog
   x = VarArray(size=nZeros, dom=range(nZeros))

   satisfy(
      x[0] == 0,

      AllDifferent(x),

      [(x[i], x[i + 1]) in comp_zeros for i in range(nZeros - 1)]
   )
\end{python}\end{boxpy}

\begin{remark}
The model, above, has not been fully checked to correspond exactly to the LPCP Contest statement. 
\end{remark}

This model involves an array of variables, a constraint \gb{Circuit} for the main variant, and a constraint \gb{AllDifferent} as well as several constraints \gb{Table} for the other variant. 
A series of 10 instances has been selected for the competition (5 instances per variant).
For generating an \x3 instance (file), you can execute for example:
\begin{command}
  python CrazyFrog -data=06.json
  python CrazyFrog -data=06.json -variant=table  
\end{command}

\subsection{Cryptanalysis}

\paragraph{Description.} From \cite{GMS_crypt}:

\begin{quote}
  Differential cryptanalysis aims at evaluating confidentiality by testing whether it is possible to find the secret key within a reasonable number of trials.
One can use Constraint Programming models to solve the chosen key differential attack against the standard block cipher AES \cite{GMS_crypt}.
In the first step of \cite{GMS_crypt}, there is a search for binary solutions.
Each unknown $\delta X_i$ is modeled with a $4 \times 4$ byte matrix, and a binary variable $\Delta X_i[j][k]$ is associated with every differential byte $\delta X_i[j][k]$.
These binary variables are equal to 0 if their associated differential bytes are equal to $0^8$, i.e.,
$\Delta X_i[j][k] = 0 \Leftrightarrow X_i[j][k] = X'_i[j][k] \Leftrightarrow \delta X_i[j][k] = 0^8$
and they are equal to 1 otherwise.
We also associate binary variables $\Delta K_i[j][k]$ and $\Delta Y_i[j][k]$ with every differential byte $\delta K_i[j][k] = K_i[j][k] \oplus K'_i[j][k]$ and $\delta Y_i[j][k] = Y_i[j][k] \oplus Y'_i[j][k]$, respectively.
The operations that transform $\delta X$ into $\delta X_r$ are translated into constraints between these binary variables.
For the first step, the goal is then to find solutions which satisfy these constraints (and solutions are called binary solutions).
\end{quote}

\paragraph{Data.}

Three integers are required to specify a specific instance: the number of rounds, the objective value $z$ and the number of bits forming the key.
The values used for generating the 2025 competition instances are:
\begin{quote}
(3, 4, 128), (3, 5, 128), (4, 9, 128), (4, 11, 128), (4, 12, 128), (5, 10, 128), \\ (5, 14, 128), (5, 15, 128), (6, 12, 128), (6, 16, 128), (6, 17, 128)
\end{quote}

\paragraph{Model.}
The \p3 model, in a file ‘Cryptanalysis.py’, used for the \x3 competition is close to (can be seen as the close translation of) the one submitted to the 2016 Minizinc challenge.
The original MZN model was proposed by David Gerault, Marine Minier, and Christine Solnon.

\begin{boxpy}\begin{python}
@\imp@

nRounds, z, nBits = data or (3, 4, 128)  # z (objective value) and number of bits in the key

m, n = 4, 4  # number of columns and rows per round
KC = nBits // 32  # number of columns per round of key schedule
NBK = KC + nRounds * m // KC  # number of variables to represent the components of the key

R, J, I, Q = range(nRounds), range(m), range(n), range(nRounds * m)  # rounds, columns, rows

V = [(q, q // m, q % m, (q - KC) // m, (q - KC) % m, (q - 1) // m, (q-1+m) % m) for q in Q]

def XOR(a, b, c, eq_ab=None, eq_ac=None, eq_bc=None):
   assert (eq_ab is None) == (eq_ac is None) == (eq_bc is None)
   if eq_ab is None:
      return a + b + c != 1  # regular xor
   else:
      return [a + b + c != 1, eq_ab == 1 - c, eq_ac == 1 - b, eq_bc == 1 - a]  # extended xor

def init_KS(q, r1, j1, r2, j2, r3, j3, i, k):
   if q < KC:  # for the first key schedule round
      if k == q:
         return key[r1][j1][i][k] == dK[r1][j1][i]
      else:
         return key[r1][j1][i][k] == 0
   if q % KC == 0:  # else, for SB positions 
      if k == (q // KC) * m + j1:
         return key[r1][j1][i][k] == dK[r3][j3][i + 1]
      else:
         return key[r1][j1][i][k] == key[r2][j2][i][k]
   return None

def aux_KS(q, r1, j1, r2, j2, r3, j3, i):
   if q % KC == 0:
      return XOR(dK[r2][j2][i], dK[r3][j3][i + 1], dK[r1][j1][i])
   return (
      XOR(
         dK[r2][j2][i], dK[r3][j3][i], dK[r1][j1][i],
         eqK[r3][j3][r2][j2][i], eqK[r1][j1][r2][j2][i], eqK[r1][j1][r3][j3][i]
      ),
      [(key[r2][j2][i][k] * dK[r2][j2][i] != key[r3][j3][i][k] * dK[r3][j3][i])
         == key[r1][j1][i][k] for k in range(NBK)]
   )
      
# dX[r][j][i] is 0 if the differential byte for X at round r, column j and row i is 0^8
dX = VarArray(size=[R, J, I], dom={0, 1})  # state after ARK (AddRoundKey)

# dY[r][j][i] is 0 if the differential byte for Y at round r, column j and row i is 0^8
dY = VarArray(size=[R[:-1], J, I], dom={0, 1})  # state before ARK

# dK[r][j][i] is 0 if the differential byte for K at round r, column j and row i is 0^8
dK = VarArray(size=[R, J, I], dom={0, 1})

# dSR[r][j][i] is the state after SR at round r, column j and row i
dSR = VarArray(size=[R, J, I], dom={0, 1})  # state after SR (ShiftRows)

# key[r][j][i][k] is the kth component of the key for round r, column j and row i
key = VarArray(size=[R, J, I, NBK], dom={0, 1})

# cX[r][j] is the sum of dX[r][][j]
cX = VarArray(size=[R, J], dom=range(m + 1))

# cK[r][j] is the sum of dK[r][][j]
cK = VarArray(size=[R, J], dom=range(m + 1))

# cSR[r][j] is the sum of dSR[r][][j]
cSR = VarArray(size=[R, J], dom=range(m + 1))

# eqK[r1][j1][r2][j2][i] is 1 if byte values of dK[r1][j1][i] and dK[r2][j2][i] are equal
eqK = VarArray(size=[R, J, R, J, I], dom={0, 1})

# eqY[q1][q2] is a lower bound on the equalities between MC(SB(A)) and MC(SB(B))
eqY = VarArray(size=[Q, Q], dom=lambda q1, q2: range(n + 1) if q1 < q2 else None)

# eqSR[q1][q2] is a lower bound on the equalities between SR(SB(A)) and SR(SB(B))
eqSR = VarArray(size=[Q, Q], dom=lambda q1, q2: range(n + 1) if q1 < q2 else None)

dff = VarArray(size=[Q, Q], dom=lambda q1, q2: {0, 1} if q1 < q2 else None)

satisfy(
   # ensuring the goal is reached
   Sum(cSR) + Sum(cK[:, KC - 1]) == z,

   # initialisation of the redundant variables
   [
      (
         cX[r][j] == Sum(dX[r][j]),
         cK[r][j] == Sum(dK[r][j]),
         cSR[r][j] == Sum(dSR[r][j])
      ) for r in R for j in J
   ],

   # operation ARK (AddRoundKey)
   [XOR(dY[r - 1][j][i], dK[r][j][i], dX[r][j][i]) for r in R[1:] for j in J for i in I],

   # ensuring Maximum Distance Separable (MDS) property of Mix Columns (MC)
   [
      [dSR[r][j][i] == dX[r][j + i][i] for r in R for j in J for i in I],
      [cSR[r][j] + Sum(dY[r][j]) in {0, 5, 6, 7, 8} for r in R[:-1] for j in J]
   ],

   # operation KS (KeySchedule)
   [
      [init_KS(*v, i, k) for v in V for i in I for k in range(NBK)],
      [aux_KS(*v, i) for v in V if v[0] >= KC for i in I],
      [Sum(key[r][j][i]) + dK[r][j][i] != 1 for r in R for j in J if r * m + j >= KC
         for i in I]
   ],

   # equality relations
   [
      (
         If(  
            eqK[r1][j1][r2][j2][i],
            Then=dK[r1][j1][i] == dK[r2][j2][i]
         ),

         eqK[r1][j1][r2][j2][i] == eqK[r2][j2][r1][j1][i],  # symmetry

         If(  
            [key[r1][j1][i][k] == key[r2][j2][i][k] for k in range(NBK)],
            Then=eqK[r1][j1][r2][j2][i] == 1
         ),

         Sum(dK[r1][j1][i], dK[r2][j2][i], eqK[r1][j1][r2][j2][i]) != 0,  

         [Sum(eqK[r1][j1][r3][j3][i], eqK[r1][j1][r2][j2][i], eqK[r2][j2][r3][j3][i]) != 2
            for r3 in R for j3 in J]  # transitivity
         
      ) for q1, q2 in combinations(Q, 2)
          if (r1 := q1 // m, j1 := q1 % m, r2 := q2 // m, j2 := q2 % m) for i in I
   ],

   # operation linear MC (MixColumns)
   [
      (
         dff[q1][q2] == Exist(
            disjunction(
               dSR[r1 - 1][j1][i] != dSR[r2 - 1][j2][i],  # different bit values
               dY[r1 - 1][j1][i] != dY[r2 - 1][j2][i],  # different after MC 
               eqK[r1][j1][r2][j2][i] + dX[r1][j1][i] + dX[r2][j2][i] == 0,  
               both(  
                  eqK[r1][j1][r2][j2][i] + dK[r1][j1][i] == 2,
                  dX[r1][j1][i] != dX[r2][j2][i]
               ) 
            ) for i in I
         ),

         eqSR[q1][q2] == Sum(dSR[r1 - 1][j1][i] + dSR[r2 - 1][j2][i] == 0 for i in I),  

         eqY[q1][q2] == Sum(  
            either(
               both(
                  eqK[r1][j1][r2][j2][i],
                  dX[r1][j1][i] + dX[r2][j2][i] == 0
               ),
               dY[r1 - 1][j1][i] + dY[r2 - 1][j2][i] == 0
            ) for i in I
         )
      ) for q1, q2 in combinations(Q[m:], 2)
          if (r1 := q1 // m, j1 := q1 % m, r2 := q2 // m, j2 := q2 % m)
   ],

   [  # If S(A) != S(B) (in bytes) then MDS Property
      If(
         dff[q1][q2],
         Then=eqSR[q1][q2] + eqY[q1][q2] <= 3
      ) for q1, q2 in combinations(Q[m:], 2)
   ]
)
\end{python}\end{boxpy}

This model involves many arrays of variables, many intensional constraints and constraints \gb{Sum}.
A series of 11 instances has been selected for the competition.
For generating an \x3 instance (file), you can execute for example:
\begin{command}
  python Cryptanalysis.py -data=[3,5,128]
\end{command}

\subsection{EFPA}

\paragraph{Description.}

From \href{https://www.csplib.org/Problems/prob055/}{CSPLib}:
\begin{quote}
The problem EFPA (Equidistant Frequency Permutation Arrays) is to find a set (optionally of maximal size) of codewords, such that any pair of $n$ codewords are Hamming distance $d$ apart.
Each codeword is made up of symbols from the alphabet $\{1 \dots,q\}$, with each symbol occurring a fixed number $\lambda$ of times per codeword.
\end{quote}

\paragraph{Data.}
Four integers are required to specify a specific instance.
The values used for generating the 2025 competition instances are:
\begin{quote}
  (3, 7, 7, 6), (3, 7, 7, 7), (3, 8, 8, 7), (3, 8, 8, 8), (4, 4, 5, 10), (4, 4, 5, 11), \\
  (4, 5, 4, 10), (4, 5, 4, 11),  (5, 4, 4, 8), (5, 4, 4, 9), (6, 4, 4, 13), (6, 4, 4, 14)
\end{quote}

\paragraph{Model.}
The \p3 model, in a file `EFPA.py', used for the competition is: 

\begin{boxpy}\begin{python}
@\imp@

d, ld, q, n = data

# x[i] is the ith symbol of the sequence
x = VarArray(size=[n, q * ld], dom=range(q))

satisfy(
   # respecting the occurrences of symbols
   [Cardinality(within=x[i], occurrences={j: ld for j in range(q)}) for i in range(n)],

   # ensuring a specific Hamming distance
   [Hamming(x[i], x[j]) == d for i, j in combinations(n, 2)],

   # tag(symmetry-breaking)
   LexIncreasing(x, matrix=True)  # strict not possible (except maybe for the rows)
)
\end{python}\end{boxpy}

This model involves an array of variables, a constraint \gb{LexMatrix}, and some constraints \gb{Cardinality} and \gb{Count}.
A series of 12 instances has been selected for the competition.
For generating an \x3 instance (file), you can execute for example:
\begin{command}
  python EFPA.py -data=[3,7,7,6]
\end{command}

\subsection{Graceful Graph}

\paragraph{Description.}

From \href{https://www.csplib.org/Problems/prob053/}{CSPLib}:
\begin{quote}
A labelling $f$ of the nodes of a graph with $q$ edges is graceful if $f$ assigns each node a unique label from $0, 1, \dots,q$
and when each edge $(i,j)$ is labelled with $|f(j) - f(j)|$, the edge labels are all different.
\end{quote}

\paragraph{Data.}
Two integers are required to specify a specific instance: a pair $(k,p)$ where $k$ is the size of each clique and $p$ is the size of each path (the number of clique).
The values used for generating the 2025 competition instances are:
\begin{quote}
(3, 8), (3, 10), (4, 5), (4, 6), (5, 4), (5, 5), (5, 6), (6, 2), (6, 3), (6, 4)
\end{quote}

\paragraph{Model.}
The \p3 model, in a file `EFPA.py', used for the competition is: 

\begin{boxpy}\begin{python}
@\imp@

k, p = data or (2, 4)  

nEdges = int(((k * (k - 1)) * p) / 2 + k * (p - 1))

# cn[i][j] is the color of the jth node of the ith clique
cn = VarArray(size=[p, k], dom=range(nEdges + 1))

# ce[i][j1][j2] is the color of the edge (j1,j2) of the ith clique
ce = VarArray(size=[p, k, k], dom=lambda i,j1,j2: range(1, nEdges + 1) if j1 < j2 else None)

# cp[i][j] is the color of the jth edge of the ith path
cp = VarArray(size=[p - 1, k], dom=range(1, nEdges + 1))

satisfy(
   # all nodes are colored differently
   AllDifferent(cn),

   # all edges are colored differently
   AllDifferent(ce + cp),

   # computing colors of edges from colors of nodes
   [
      [ce[i][j1][j2] == abs(cn[i][j1] - cn[i][j2])
         for i in range(p) for j1, j2 in combinations(k, 2)],

      [cp[i][j] == abs(cn[i][j] - cn[i + 1][j])
         for i in range(p - 1) for j in range(k)]
   ]
)
\end{python}\end{boxpy}

This model involves an array of variables, and two global constraints \gb{AllDifferent}.
A series of 10 instances has been selected for the competition.
For generating an \x3 instance (file), you can execute for example:
\begin{command}
  python GracefulGraph.py -data=[3,8]
\end{command}

\subsection{Heterosquare}

\paragraph{Description.}

From \href{https://mathworld.wolfram.com/Heterosquare.html}{mathworld.wolfram.com}:
\begin{quote}
A heterosquare is an $n \times n$ array of the integers from 1 to $n^2$ such that the rows, columns, and diagonals have different sums.
By contrast, in a magic square, they have the same sum.
There are no heterosquares of order two, but heterosquares of every odd order exist.
They can be constructed by placing consecutive integers in a spiral pattern.
\end{quote}

\paragraph{Data.}
An integer is required to specify a specific instance: the order $n$ of the problem instance. 
The values used for generating the 2025 competition instances are:
\begin{quote}
  20, 25, 30, 35, 40, 50, 60 \hfill for variant "easy" \\
  5, 10, 15, 20 \hfill for variants "fair" and "hard"
\end{quote}

\paragraph{Model.}
The \p3 model, in a file `Heterosquare.py', used for the competition is: 

\begin{boxpy}\begin{python}
@\imp@

assert not variant() or  variant() in ("easy", "fair", "hard")

n = data or 8

lb, ub = (n * (n + 1)) // 2, ((n * n) * (n * n + 1)) // 2

if variant():
   if variant("easy"):
      lb = lb * (n // 2)
   elif variant("fair"):
      lb = lb * (n - 1)
   elif variant("hard"):
      lb = lb * n
   ub = ub // (n // 2)

# x[i][j] is the value put in cell of the matrix at coordinates (i,j)
x = VarArray(size=[n, n], dom=range(1, n * n + 1))

# rs[i] is the sum of values in the ith row
rs = VarArray(size=n, dom=range(lb, ub + 1))

# cs[j] is the sum of values in the jth column
cs = VarArray(size=n, dom=range(lb, ub + 1))

# ds is the sum in the two diagonals
ds = VarArray(size=2, dom=range(lb, ub + 1))


satisfy(
   # all values must be different
   AllDifferent(x),

   # computing row sums
   [rs[i] == Sum(x[i]) for i in range(n)],

   # computing column sums
   [cs[j] == Sum(x[:, j]) for j in range(n)],

   # computing diagonal sums
   [
      ds[0] == Sum(diagonal_down(x)),
      ds[1] == Sum(diagonal_up(x))
   ],

   # all sums must be different
   AllDifferent(rs, cs, ds),

   # ensuring Frenicle standard form  tag(symmetry-breaking)
   [
      x[0][0] < x[0][-1],
      x[0][0] < x[-1][0],
      x[0][0] < x[-1][-1],
      x[0][1] < x[1][0]
   ]
)
\end{python}\end{boxpy}

This model involves four arrays of variables, and global constraints \gb{AllDifferent} and \gb{Sum}.
A series of 15 instances (7 easy instances, 4 fair instances and 4 hard instances) has been selected for the competition.
For generating an \x3 instance (file), you can execute for example:
\begin{command}
  python Heterosquare.py -data=20 -variant=easy
  python Heterosquare.py -data=20 -variant=fair
  python Heterosquare.py -data=20 -variant=hard
\end{command}

\subsection{LangfordBin}

\paragraph{Description.}

From \cite{GJM_watched}: 
\begin{quote}
Langford’s Problem $L(k, n)$ requires finding a list of length $k \times n$, which contains $k$ sets of the numbers 1 to $n$,
such that for all $m \in \{1, 2, \dots, n\}$ there is a gap of size $m$ between adjacent occurrences of $m$.
In the constraint solver Minion \cite{GJM_minion}, $L(2,n)$ was modelled using two vectors of variables, $V$ and $P$, each of size $2n$.
Each variable in $V$, which represnts the result, has domain $\{1, 2, \dots n\}$.
For each $i \in \{1, 2, \dots\}$ the $2i$th and $2i + 1$st variables in $P$ are the first and second positions of $i$ in $V$.
Each variable in $P$ has domain $\{0, 1, \dots, 2n - 1\}$, indexing matrices from 0.
\end{quote}

\paragraph{Data.}
An integer is required to specify a specific instance: the order $n$ of the problem instance. 
The values used for generating the 2025 competition instances are:
\begin{quote}
10, 13, 14, 16, 39, 40, 59, 60, 79, 80, 120, 240
\end{quote}

\paragraph{Model.}
The \p3 model, in a file `LangfordBin.py', used for the competition is: 

\begin{boxpy}\begin{python}
@\imp@

n = data or 8
N = range(n)

# v[i] is the ith value of the Langford's sequence
v = VarArray(size=2 * n, dom=range(1, n + 1))

# p[j] is the first (resp., second) position of 1+j/2 in v if j is even (resp., odd)
p = VarArray(size=2 * n, dom=range(2 * n))

satisfy(
    [v[p[2 * i]] == i + 1 for i in N],

    [v[p[2 * i + 1]] == i + 1 for i in N],

    [p[2 * i] == i + 2 + p[2 * i + 1] for i in N]
)
\end{python}\end{boxpy}

This model involves two arrays of variables, and global constraints \gb{Element}.
A series of 12 instances has been selected for the competition.
For generating an \x3 instance (file), you can execute for example:
\begin{command}
  python LangfordBin.py -data=14
\end{command}

\subsection{Lottery Design}

\paragraph{Description.}

See \cite{CS_lottery}:
\begin{quote}
$X$ is a set of balls labelled 1 to $d$.
Find a set $B$ of $n$ tickets each containing $m$ of these numbers, such that for any draw $D$ of $p$ distinct
balls from $X$, we can find at least one ticket which matches $D$ in at least $t$ places.
\end{quote}

\paragraph{Data.}
Five integers are required to specify a specific instance.
The values used for generating the 2025 competition instances are:
\begin{quote}
$(32, 6, 6, 2, k)$ with $k$ in $\{7, 17, 47, 67, 87, 107, 207, 307, 407, 1000, 2000, 3000\}$
\end{quote}

\paragraph{Model.}
The \p3 model, in a file `LotteryDesign.py', used for the competition is: 

\begin{boxpy}\begin{python}
@\imp@

d, m, p, t, n = data or (32, 6, 6, 2, 17)

assert t == 2  # for the moment

# x[i][j] is the jth value on the ith ticket
x = VarArray(size=[n, m], dom=range(d))

# draw[k] is the kth value of a draw
draw = VarArray(size=p, dom=range(d))

tol = VarArray(size=n, dom=range(p))


satisfy(
   AllDifferentList(x),

   # tag(symmetry-breaking)
   [Increasing(x[i], strict=True) for i in range(n)],

   Increasing(draw, strict=True),

   [
      If(
         tol[i] != k,
         Then=x[i][j] != draw[k]
      ) for i in range(n) for j in range(m) for k in range(p)
   ]
)
\end{python}\end{boxpy}

\begin{remark}
The model, above, does not exactly to the statement of the original problem (it was written for the 2025 XCSP3 competition).
\end{remark}

This model involves three arrays of variables, and the global constraint \gb{AllDifferentList}.
A series of 12 instances has been selected for the competition.
For generating an \x3 instance (file), you can execute for example:
\begin{command}
  python LotteryDesign.py -data=[32,6,6,2,17]
\end{command}

\subsection{PegSolitaireTable}

\paragraph{Description.}

Peg Solitaire is played on a board with a number of holes.
In the English version of the game considered here, the board is in the shape of a cross with 33 holes.
Pegs are arranged on the board so that at least one hole remains. By making horizontal or vertical draughts-like moves,
the pegs are gradually removed until a goal configuration is obtained.
In the classic ‘central’ Solitaire, the goal is to reverse the starting position, leaving just a single peg in the central hole (e.g., see \cite{JMMT_modelling}).

\paragraph{Data.}
Three integers are required to specify a specific instance: the coordinates of the starting point, and the number of allowed moves.
The values used for generating the 2025 competition instances are:
\begin{quote}
(0, 2, 0), (0, 3, 0), (0, 4, 0), (1, 2, 0), (1, 3, 0), (1, 4, 0), (2, 0, 0), (2, 2, 0), (2, 3, 0), (2, 4, 0), (2, 6, 0), (3, 3, 0)
\end{quote}

\paragraph{Model.}
The \p3 model, in a file `PegSolitaireTable.py', used for the competition is: 

\begin{boxpy}\begin{python}
@\imp@

from data.PegSolitaire_Generator import generate_boards, build_transitions

assert variant() in {"english", "french"}

origin_x, origin_y, nMoves = data or (0, 2, 0)

init_board, final_board = generate_boards(variant(), origin_x, origin_y)
n, m = len(init_board), len(init_board[0])
transitions = build_transitions(init_board)
nTransitions = len(transitions)

horizon = sum(sum(v for v in row if v) for row in init_board)
           - sum(sum(v for v in row if v) for row in final_board)
nMoves = horizon if nMoves <= 0 or horizon < nMoves else nMoves
assert 0 < nMoves <= horizon

P = [(i, j) for i in range(n) for j in range(m) if init_board[i][j] is not None]

def peg_table(i, j, t):
   V1 = [k for k, tr in enumerate(transitions) if (i, j) in (tr[0:2], tr[2:4], tr[4:6])]
   V2 = [k for k, tr in enumerate(transitions) if (i, j) in (tr[0:2], tr[2:4])]
   V3 = [k for k, tr in enumerate(transitions) if (i, j) == tr[4:6]]
   T = []
   for k in range(nTransitions):
      for v1 in (0, 1):
         for v2 in (0, 1):
            if not ((k not in V1) == (v1 == v2)):
               continue
            if not ((k in V2) == (v1 == 1 and v2 == 0)):
               continue
            if not ((k in V3) == (v1 == 0 and v2 == 1)):
               continue
            T.append((k, v1, v2))
   return T


# x[i][j][t] is the value at row i and column j at time t
x = VarArray(
   size=[nMoves+1, n, m],
   dom=lambda t, i, j: {0, 1} if init_board[i][j] is not None else None
)

# y[t] is the move (transition) performed at time t
y = VarArray(size=nMoves, dom=range(nTransitions))

satisfy(
   # setting the initial board
   x[0] == init_board,

   # setting the final board
   x[-1] == final_board,

   # handling unchanged situations
   [
     Table(
        scope=(y[t], x[t][i][j], x[t + 1][i][j]),
        supports=peg_table(i, j, t)
     ) for (i, j) in P for t in range(nMoves)
   ]
)
\end{python}\end{boxpy}

This model involves two arrays of variables, and constraints \gb{Table}.
A series of 12 instances has been selected for the competition.
For generating an \x3 instance (file), you can execute for example:
\begin{command}
  python PegSolitaireTable.py -data=[0,2,0]
\end{command}

\subsection{Ramsey Partition}

\paragraph{Description.}

From \href{http://www.mathematik.uni-bielefeld.de/~sillke/PUZZLES/partion3-ramsey}{mathematik.uni-bielefeld.de}:
\begin{quote}
Partition the integers 1 to $n$ into three sets, such that for no set are
there three different numbers with two adding to the third.
\end{quote}

\paragraph{Data.}
Two integers are required to specify a specific instance; the number $q$ of sets and the number of values $n$.
The values used for generating the 2025 competition instances are:
\begin{quote}
  (3, 300), (3, 600), (3, 1200), (4, 60), (4, 62), (4, 64), (4, 80), (4, 100), (4, 400), \\
  (5, 130), (5, 140), (5, 160), (5, 180), (5, 200),  (5, 400)
\end{quote}

\paragraph{Model.}
The \p3 model, in a file `RamseyPartition.py', used for the competition is: 

\begin{boxpy}\begin{python}
@\imp@

q, n, = data or (3, 23)

L = n // q 

Q = range(q)

# x[i] is the set (index) where is put the value (i+1)
x = VarArray(size=n, dom=range(q))

satisfy(
   # ensuring some occurrences are guaranteed
   Cardinality(
      within=x,
      occurrences={v: L for v in Q}
   ),

   # ensuring no three different numbers with two adding to the third in teh same set
   [
     NValues(x[i],x[j],x[k]) > 1
        for i, j, k in combinations(n, 3) if i + 1 + j + 1 == k + 1
   ]
)
\end{python}\end{boxpy}

This model involves an array of variables, and the global constraints \gb{Cardinality} and \gb{NValues}.
A series of 15 instances has been selected for the competition.
For generating an \x3 instance (file), you can execute for example:
\begin{command}
  python RamseyPartition.py -data=[3,23]
\end{command}

\subsection{Rostering}

\paragraph{Description.}

From \cite{PQZ_counting}: 

\begin{quote}
This problem was inspired by a rostering context.
The objective is to schedule n employees over a span of $n$ time periods.
In each time period, $n-1$ tasks need to be accomplished and one employee out of the $n$ has a break.
The tasks are fully ordered 1 to $n-1$; for each employee the schedule has to respect the following rules:
\begin{itemize}
\item two consecutive time periods have to be assigned to either two consecutive tasks, in no matter which order i.e. $(t, t+1)$ or $(t+1, t)$,
    or to the same task i.e. $(t, t)$;
\item an employee can have a break after no matter which task;
 \item  after a break an employee cannot perform the task that precedes the task prior to the break, i.e. ($t$, break, $t-1$) is not allowed.
\end{itemize}
   The problem is modeled with one constraint \gb{Regular} per row and one constraint \gb{Alldifferent} per column.
\end{quote}

\paragraph{Data.}

As an illustration of data specifying an instance of this problem, we have:
\begin{json}
{
  "preset": [
    [7, 5, 8],
    [7, 0, 8],
    [3, 5, 5],
    [0, 5, 4],
    [1, 7, 7]
  ],
  "forbidden": []
}
\end{json}

\paragraph{Model.}

The \p3 model, in a file `Rostering.py', used for the \x3 competition is:

\begin{boxpy}\begin{python}
@\imp@

preset, forbidden, n = data or load_json_data("roster-5-00-02.json")

def automaton():  # NB: some rules above are made stricter than JAIR paper's rules
   # q(1,i) means before break and just after reading i
   # q(2,i) means just after reading break (0) and i before
   # q(3,i) means after break and just after reading i
   q, N = Automaton.q, range(1, n)
   t = [(q(0), 0, q(2, 0))] + [(q(2, 0), i, q(3, i)) for i in N]
   t.extend((q(0), i, q(1, i)) for i in N)
   t.extend((q(1, i), j, q(1, j)) for i in N for j in (i - 1, i + 1) if 1 <= j < n)
   t.extend((q(1, i), 0, q(2, i)) for i in N)
   t.extend((q(2, i), j, q(3, j)) for i in N for j in N if abs(i - j) != 1)
   t.extend((q(3, i), j, q(3, j)) for i in N for j in (i - 1, i + 1) if 1 <= j < n)
   return Automaton(
      start=q(0),
      final=[q(2, i) for i in N] + [q(3, i) for i in N],
      transitions=t
   )

A = automaton()

# x[i][j] is the task (or break) performed by the ith employee at time j
x = VarArray(size=[n, n], dom=range(n))

satisfy(
   # respecting preset tasks
   [x[i][j] == k for (i, j, k) in preset],

   # respecting forbidden assignments
   [x[i][j] != k for (i, j, k) in forbidden],

   # respecting job rules for each employee
   [x[i] in A for i in range(n)],

   # all tasks are different at any time
   [AllDifferent(x[:, j]) for j in range(n)]
)
\end{python}\end{boxpy}

\begin{remark}
The model/automaton above is made stricter (wrt \cite{PQZ_counting}) so as (hopefully) to generate harder instances.
\end{remark}

This model involves an array of variables and (global) constraints of type \gb{Regular} and \gb{AllDifferent}.

A series of 15 instances has been selected for the competition.
For generating an \x3 instance (file), you can execute for example:
\begin{command}
python Rostering.py -data=inst.json
\end{command}
where `inst.json' is a data file in JSON format.

\subsection{Rotating Rostering}

\paragraph{Description.}

From \href{https://www.csplib.org/Problems/prob087/}{CSPLib}:

\begin{quote}
This problem is taken from real life rostering challenges (like nurse rostering).
The task is it to find a shift assignment for every employee for every day.
A rotation system is used to decrease the size of the problem.
Thus, only the rostering for one employee is calculated and all other employees gain a rotated version of the rostering.
So Employee 2 has in the first week the rostering of Employee 1 in the second week.
Employee 3 has in the first week the rostering of Employee 2 in the second week and Employee 1 in the third week etc.
\end{quote}

\paragraph{Data.}

As an illustration of data specifying an instance of this problem, we have:
\begin{json}
{
  "nDaysPerWeek": 7,
  "nWeeks": 8,
  "shift_min": 2,
  "shift_max": 3,
  "requirements": [
    [2, 2, 2, 2],
    [2, 2, 2, 2],
    [2, 2, 2, 2],
    [2, 2, 2, 2],
    [2, 2, 2, 2],
    [4, 2, 1, 1],
    [4, 2, 1, 1]
  ]
}
\end{json}

\paragraph{Model.}

The \p3 model, in a file `RotatingRostering.py', used for the \x3 competition is:

\begin{boxpy}\begin{python}
@\imp@

nDaysPerWeek, nWeeks, shift_min, shift_max, requirements = data 

SATURDAY, SUNDAY = 5, 6
shifts = (OFF, EARLY, LATE, NIGHT) = range(4)
nShifts = len(shifts)

nDays = nWeeks * nDaysPerWeek
D, W = range(nDays), range(nWeeks)

# x[i] is the shift for the ith day (along the flattened horizon)
x = VarArray(size=nDays, dom=range(nShifts))

# y[w][d] is the shift in the dth day of the wth week
y = VarArray(size=[nWeeks, nDaysPerWeek], dom=range(nShifts))

satisfy(
   # computing y
   [y[w][d] == x[w * nDaysPerWeek + d] for w in W for d in range(nDaysPerWeek)],

   # ensuring weekend days (Saturday and Sunday) have the same shift
   [y[w][SATURDAY] == y[w][SUNDAY] for w in W],
   
   # ensuring a minimum length of consecutive similar shifts
   [
      If(
         x[i] != x[i + 1],
         Then=AllEqual(x[i + 1:i + 1 + shift_min])
      ) for i in D
   ],

   # ensuring a maximum length of consecutive similar shifts
   [
      If(
         AllEqual(x[i:i + shift_max]),
         Then=x[i] != x[i + shift_max]
      ) for i in D
   ],

   # ensuring to have at least 2 resting days every 2 weeks
   [
      Count(
         within=x[i:i + 2 * nDaysPerWeek + 1],
         value=OFF
      ) >= 2 for i in D  # does seem to be +1 in the model posted at CSPLib?
   ],

   # avoiding some specific successive shifts
   [
      Table(
         scope=(x[i], x[i + 1]),
         conflicts={(LATE, EARLY), (NIGHT, EARLY), (NIGHT, LATE)}
      ) for i in D
   ],

   # ensuring the right number of employees
   [
      Cardinality(
         within=y[:, d],
         occurrences={s: requirements[d][s] for s in range(nShifts)}
      ) for d in range(nDaysPerWeek)
   ]
)
\end{python}\end{boxpy}

This model involves two arrays of variables and (global) constraints of type \gb{AllEqual}, \gb{Cardinality}, \gb{Count} and \gb{Table}.

A series of 14 instances has been selected for the competition.
For generating an \x3 instance (file), you can execute for example:
\begin{command}
python RotatingRostering.py -data=inst.json
\end{command}
where `inst.json' is a data file in JSON format.

\subsection{SEDF}

\paragraph{Description.}

From \cite{tabID}:

\begin{quote}
A Strong External Difference Family (SEDF) is an object defined on a group, with applications in communications and cryptography.
Given a finite group $G$ on a set of size $n$, an $(n,m,k,\lambda)$ SEDF is a list $A_1, \dots, A_m$ of disjoint subsets of size $k$ of $G$ such that,
for all $1 \leq i \leq m$, the multi-set $\{xy-1 \mid x \in A_i , y \in A_j, i \neq j\}$ contains $\lambda$ occurrences of each non-identity element of $G$.
The parameters of the SEDF problem are $(n,m,k,\lambda)$, the group $G$ given as a multiplication table \texttt{tab} (which is an $n \times n$ matrix of integers),
and \texttt{inv}, a one-dimensional table which maps each group element to its inverse.
The SEDF is represented as an $m \times k$ matrix.
\end{quote}

\paragraph{Data.}

As an illustration of data specifying an instance of this problem, we have:
\begin{json}
{
  "inv": [1, 19, 18, 17, 16, 15, 14, 13, 12, 11, 10, 9, 8, 7, 6, 5, 4, 3, 2],
  "tab": [
    [1, 2, 3, 4, 5, 6, 7, 8, 9, 10, 11, 12, 13, 14, 15, 16, 17, 18, 19],
    [2, 3, 4, 5, 6, 7, 8, 9, 10, 11, 12, 13, 14, 15, 16, 17, 18, 19, 1],
    [3, 4, 5, 6, 7, 8, 9, 10, 11, 12, 13, 14, 15, 16, 17, 18, 19, 1, 2],
    [4, 5, 6, 7, 8, 9, 10, 11, 12, 13, 14, 15, 16, 17, 18, 19, 1, 2, 3],
    [5, 6, 7, 8, 9, 10, 11, 12, 13, 14, 15, 16, 17, 18, 19, 1, 2, 3, 4],
    [6, 7, 8, 9, 10, 11, 12, 13, 14, 15, 16, 17, 18, 19, 1, 2, 3, 4, 5],
    [7, 8, 9, 10, 11, 12, 13, 14, 15, 16, 17, 18, 19, 1, 2, 3, 4, 5, 6],
    [8, 9, 10, 11, 12, 13, 14, 15, 16, 17, 18, 19, 1, 2, 3, 4, 5, 6, 7],
    [9, 10, 11, 12, 13, 14, 15, 16, 17, 18, 19, 1, 2, 3, 4, 5, 6, 7, 8],
    [10, 11, 12, 13, 14, 15, 16, 17, 18, 19, 1, 2, 3, 4, 5, 6, 7, 8, 9],
    [11, 12, 13, 14, 15, 16, 17, 18, 19, 1, 2, 3, 4, 5, 6, 7, 8, 9, 10],
    [12, 13, 14, 15, 16, 17, 18, 19, 1, 2, 3, 4, 5, 6, 7, 8, 9, 10, 11],
    [13, 14, 15, 16, 17, 18, 19, 1, 2, 3, 4, 5, 6, 7, 8, 9, 10, 11, 12],
    [14, 15, 16, 17, 18, 19, 1, 2, 3, 4, 5, 6, 7, 8, 9, 10, 11, 12, 13],
    [15, 16, 17, 18, 19, 1, 2, 3, 4, 5, 6, 7, 8, 9, 10, 11, 12, 13, 14],
    [16, 17, 18, 19, 1, 2, 3, 4, 5, 6, 7, 8, 9, 10, 11, 12, 13, 14, 15],
    [17, 18, 19, 1, 2, 3, 4, 5, 6, 7, 8, 9, 10, 11, 12, 13, 14, 15, 16],
    [18, 19, 1, 2, 3, 4, 5, 6, 7, 8, 9, 10, 11, 12, 13, 14, 15, 16, 17],
    [19, 1, 2, 3, 4, 5, 6, 7, 8, 9, 10, 11, 12, 13, 14, 15, 16, 17, 18]
  ],
  "k": 3,
  "m": 3,
  "ld": 1
}
\end{json}

\paragraph{Model.}

The \p3 model, in a file `SEDF.py', used for the \x3 competition is:

\begin{boxpy}\begin{python}
@\imp@

inv, tab, k, m, ld = data or load_json_data("19-1-C19-3-3-1.json")

n = len(inv)

N, M, K = range(n), range(m), range(k)

# x[i][j] is the jth value of the ith list
x = VarArray(size=[m, k], dom=range(1, n + 1))

satisfy(
   # ensuring all different values
   AllDifferent(x),

   [Increasing(x[i], strict=True) for i in M],

   Increasing(x[:, 0], strict=True),

   # ensuring the right number of occurrences
   [
      Cardinality(
         within=[tab[x[i1][j1], inv[x[i2][j2]]]
                    for j1 in K for i2 in M if i1 != i2 for j2 in K],
         occurrences={v + 1: 0 if v == 0 else ld for v in N}
      ) for i1 in M
   ]
)
\end{python}\end{boxpy}

This model involves an array of variables and (global) constraints of type \gb{AllDifferent}, \gb{Cardinality} and \gb{Element}.

A series of 10 instances has been selected for the competition.
For generating an \x3 instance (file), you can execute for example:
\begin{command}
python SEDF.py -data=inst.json
\end{command}
where `inst.json' is a data file in JSON format.

\subsection{Tiling Rythmic Canons}

\paragraph{Description.}

For example, see \cite{sat_tiling}:

\begin{quote}
Given a period $n$ and a rhythm $A$ subset of $Z_n$, the Aperiodic Tiling Complements Problem consists in finding all its aperiodic complements $B$,
i.e., all subsets $B$ of $Z_n$ such that $A \oplus B = Z_n$.  
\end{quote}

\paragraph{Data.}

As an illustration of data specifying an instance of this problem, we have:
\begin{json}
{
  "n": 168,
  "D": [24, 56, 84],
  "A": [0, 8, 16, 24, 32, 40, 42, 48, 50, 58, 66, 74, 82, 90]
}
\end{json}

\paragraph{Model.}

The \p3 model, in a file `TilingRythmicCanons.py', used for the \x3 competition is:

\begin{boxpy}\begin{python}
@\imp@

n, D, A = data

lD, lA, lB = len(D), len(A), n // len(A)

# x[i] is the ith value of the aperiodic tiling complement
x = VarArray(size=lB, dom=range(n))

# xc[k][i] is the ith value of the kth tiling complement equivalent to x under translation 
# tag(symmetry-breaking)
xc = VarArray(size=[lB - 1, lB], dom=range(n))

satisfy(
   # starting with 0
   x[0] == 0,

   # ordering values of the tiling complement
   Increasing(x, strict=True),

   # tag(symmetry-breaking)
   [
      (
         xc[i][0] == n - x[i + 1],  # gap in the first column
         xc[i][i + 1] == 0,  # 0 in the diagonal
         [xc[i][j] == (x[j] + xc[i][0]) % n for j in range(1, lB) if j != i + 1],
         LexIncreasing(x, xc[i][i + 1: i + 1 + lB], strict=True)  
      ) for i in range(lB - 1) 
   ],

   # ensuring a tiling rhythmic canon with period 'n'
   AllDifferent((A[i] + x[j]) % n for i in range(lA) for j in range(lB)),

   # ensuring the complement is aperiodic  tag(aperiodicity)
   [NValues(x, [(d + x[j]) % n for j in range(lB)]) > lB for d in D]
)
\end{python}\end{boxpy}

This model involves two arrays of variables and (global) constraints of type \gb{AllDifferent}, \gb{Lex} and \gb{NValues}.

A series of 14 instances has been selected for the competition.
For generating an \x3 instance (file), you can execute for example:
\begin{command}
python TilingRythmicCanons.py -data=inst.json
\end{command}
where `inst.json' is a data file in JSON format.

\section{COP}

\subsection{Altered States}

\paragraph{Description.}

From \href{https://www.janestreet.com/puzzles/altered-states-index/}{JaneStreet (pb1)} and \href{https://www.janestreet.com/puzzles/altered-states-2-index/}{JaneStreet (pb2)}:
\begin{quote}
Enter letters into a $n \times n$ grid above to achieve the highest score you can.
You earn points for each of the 50 U.S. states present in your grid.
Note that:
\begin{itemize}
\item states can be spelled by making King’s moves from square to square,
\item the score for a state is its length (main variant),
\item if a state appears multiple times in your grid, it only scores once.
\end{itemize}
\end{quote}

\paragraph{Data.}
An integer is required to specify a specific instance: the order of the grid.
The values used for generating the 2025 competition instances are:
\begin{quote}
2, 3, 4, 5, 6, 8, 10, 12, 15
\end{quote}

\paragraph{Model.}
The \p3 model, in a file `AlteredStates.py', used for the competition is: 

\begin{boxpy}\begin{python}
@\imp@
from pycsp3.classes.auxiliary.enums import TypeSquareSymmetry

assert not variant() or variant("bis")

n = data or 5

states = [
   ("Alabama", 5_024_279),  # 0
   ("Alaska", 733_391),  # 1
   ("Arizona", 7_151_502),  # 2
   ("Arkansas", 3_011_524),  # 3
   ("California", 39_538_223),  # 4
   ("Colorado", 5_773_714),  # 5
   ("Connecticut", 3_605_944),  # 6
   ("Delaware", 989_948),  # 7
   ("Florida", 21_538_187),  # 8
   ("Georgia", 10_711_908),  # 9
   ("Hawaii", 1_455_271),  # 10
   ("Idaho", 1_839_106),  # 11
   ("Illinois", 12_812_508),  # 12
   ("Indiana", 6_785_528),  # 13
   ("Iowa", 3_190_369),  # 14
   ("Kansas", 2_937_880),  # 15
   ("Kentucky", 4_505_836),  # 16
   ("Louisiana", 4_657_757),  # 17
   ("Maine", 1_362_359),  # 18
   ("Maryland", 6_177_224),  # 19
   ("Massachusetts", 7_029_917),  # 20
   ("Michigan", 10_077_331),  # 21
   ("Minnesota", 5_706_494),  # 22
   ("Mississippi", 2_961_279),  # 23
   ("Missouri", 6_154_913),  # 24
   ("Montana", 1_084_225),  # 25
   ("Nebraska", 1_961_504),  # 26
   ("Nevada", 3_104_614),  # 27
   ("NewHampshire", 1_377_529),  # 28
   ("NewJersey", 9_288_994),  # 29
   ("NewMexico", 2_117_522),  # 30
   ("NewYork", 20_201_249),  # 31
   ("NorthCarolina", 10_439_388),  # 32
   ("NorthDakota", 779_094),  # 33
   ("Ohio", 11_799_448),  # 34
   ("Oklahoma", 3_959_353),  # 35
   ("Oregon", 4_237_256),  # 36
   ("Pennsylvania", 13_002_700),  # 37
   ("RhodeIsland", 1_097_379),  # 38
   ("SouthCarolina", 5_118_425),  # 39
   ("SouthDakota", 886_667),  # 40
   ("Tennessee", 6_910_840),  # 41
   ("Texas", 29_145_505),  # 42
   ("Utah", 3_271_616),  # 43
   ("Vermont", 643_077),  # 44
   ("Virginia", 8_631_393),  # 45
   ("Washington", 7_705_281),  # 46
   ("WestVirginia", 1_793_716),  # 47
   ("Wisconsin", 5_893_718),  # 48
   ("Wyoming", 576_851)  # 49
]

names, populations = zip(*states)
lengths = [len(name) for name in names]
nStates, M = len(states), max(lengths)

symmetries = [[i * n + j for i, j in S] for sym in TypeSquareSymmetry
                if (S := flatten(sym.apply_on(n), keep_tuples=True))]

words = [alphabet_positions(name.lower()) for name in names]  # words converted to numbers 

T = [(i * n + j, (i + k) * n + (j + p))
       for i in range(n) for j in range(n)
          for k in [-1, 0, 1] for p in [-1, 0, 1]
             if 0 <= i + k < n and 0 <= j + p < n and (k, p) != (0, 0)]

# x[ij] is the letter (index) in cell whose index is ij
x = VarArray(size=n * n, dom=range(26))

# y[k] is 1 if the kth state is present in the matrix
y = VarArray(size=nStates, dom={0, 1})

# z[k][q] is the (flat) cell index in x of the qth letter of the kth state, or 0 
z = VarArray(size=[nStates, M], dom=lambda k, q: range(n * n) if q < lengths[k] else None)

satisfy(
   # ensuring the coherence of putting or not the states in the grid
   [
      If(
         y[k] == 0,
         Then=[z[k][q] == 0 for q in Q],  # we force 0 to avoid symmetries  z[k] == 0,
         Else=[
            [x[z[k][q]] == words[k][q] for q in Q] if not variant()
            else Sum(x[z[k][q]] != words[k][q] for q in Q) <= 1,  # the name must be  present
            [(z[k][q], z[k][q + 1]) in T for q in Q[:-1]]  # ensuring connectedness of letters
         ]
      ) for k in range(nStates) if (Q := range(lengths[k]))
   ],

   # tag(symmetry-breaking)
   [x <= x[symmetry] for symmetry in symmetries]
)

if not variant():
   maximize(
      y * lengths
   )
elif variant("bis"):
   maximize(
      y * populations
   )
\end{python}\end{boxpy}

This model involves three arrays of variables, and the global constraints \gb{Element}, \gb{Lex} and \gb{Sum}.
A series of 18 instances (9 instances for the main variant, and 9 instances for the variant ``bis'') has been selected for the competition.
For generating an \x3 instance (file), you can execute for example:
\begin{command}
  python AlteredStates.py -data=6
  python AlteredStates.py -data=6 -variant=bis
\end{command}

\subsection{Block Modeling}

\paragraph{Description.}

For example, see \cite{MDNS_generic}.

\begin{quote}
Block modeling has a long history in the analysis of social networks.
The core problem is to take a graph and divide it into $k$ clusters and interactions between those clusters described by a $k \times k$ image matrix.
\end{quote}

\paragraph{Data.}

As an illustration of data specifying an instance of this problem, we have:
\begin{json}
{
  "matrix": [
    [0, 1, 1, 0, 1, 1, 1, 0, 1, 0, 1, 0, 0, 0, 1, 1, 1, 0, 0, 1],
    [1, 0, 1, 0, 1, 1, 1, 0, 1, 0, 0, 0, 0, 1, 0, 1, 1, 0, 0, 0],
    [1, 0, 0, 1, 1, 1, 0, 0, 1, 0, 1, 0, 0, 0, 0, 0, 1, 0, 0, 0],
    [1, 1, 1, 0, 1, 1, 1, 0, 1, 0, 0, 0, 0, 1, 0, 0, 0, 0, 0, 0],
    [1, 0, 1, 1, 0, 1, 1, 0, 0, 0, 1, 1, 1, 1, 0, 1, 1, 0, 0, 1],
    [1, 0, 1, 1, 1, 0, 1, 0, 1, 0, 1, 1, 0, 1, 0, 0, 0, 0, 0, 0],
    [1, 0, 1, 1, 1, 1, 0, 0, 1, 1, 0, 0, 0, 0, 0, 1, 0, 0, 0, 0],
    [1, 1, 0, 0, 1, 0, 0, 0, 0, 1, 0, 0, 0, 0, 0, 0, 0, 0, 0, 0],
    [1, 0, 1, 0, 1, 1, 1, 0, 0, 0, 1, 0, 0, 0, 0, 0, 1, 0, 0, 0],
    [1, 0, 0, 0, 1, 0, 0, 0, 0, 0, 0, 0, 0, 0, 0, 0, 0, 0, 0, 0],
    [1, 0, 1, 0, 1, 1, 1, 0, 1, 0, 0, 1, 0, 0, 0, 0, 0, 0, 0, 0],
    [1, 0, 0, 0, 1, 0, 1, 0, 0, 0, 0, 0, 0, 0, 0, 0, 1, 0, 0, 0],
    [1, 0, 0, 0, 1, 0, 0, 0, 0, 0, 0, 0, 0, 0, 0, 0, 0, 0, 0, 0],
    [1, 0, 1, 0, 1, 1, 1, 0, 1, 0, 0, 1, 1, 0, 0, 0, 1, 1, 0, 1],
    [1, 0, 1, 1, 1, 1, 1, 0, 1, 1, 1, 1, 1, 1, 0, 1, 1, 1, 0, 1],
    [1, 1, 0, 0, 1, 1, 1, 0, 0, 0, 1, 0, 0, 0, 0, 0, 0, 0, 0, 0],
    [1, 1, 1, 0, 1, 1, 1, 0, 1, 1, 0, 0, 0, 1, 0, 0, 0, 0, 0, 0],
    [1, 0, 0, 0, 1, 0, 0, 0, 0, 0, 0, 0, 1, 1, 0, 0, 0, 0, 0, 0],
    [1, 1, 0, 0, 1, 1, 1, 0, 1, 0, 0, 0, 0, 0, 0, 0, 1, 0, 0, 0],
    [1, 0, 0, 0, 1, 0, 0, 0, 0, 0, 0, 0, 0, 0, 1, 1, 1, 0, 0, 0]
  ],
  "k": 2
}
\end{json}

\paragraph{Model.}

The \p3 model, in a file `BlockModeling.py', used for the \x3 competition is:

\begin{boxpy}\begin{python}
@\imp@

matrix, nBlocks = data or load_json_data("kansas-2.json")

nNodes = len(matrix)
N, B = range(nNodes), range(nBlocks)

# x[i][j] is the value in the grid at the ith row and jth column
x = VarArray(size=[nBlocks, nBlocks], dom={0, 1})

# y[k] is the block index of the kth node
y = VarArray(size=nNodes, dom=range(nBlocks))

satisfy(
   # partial symmetry breaking constraint
   y[u] <= u for u in B[:-1]
)

minimize(
   Sum(
      conjunction(
         y[i] == u,
         y[j] == v,
         x[u][v] != matrix[i][j]
      ) for u in B for v in B for i in N for j in N if i != j
   )
   + Sum(
      both(
         y[i] == u,
         x[u][u] != matrix[i][i]
      ) for u in B for i in N
   )
)
\end{python}\end{boxpy}

This model involves two arrays of variables and (global) constraints of type \gb{Sum}.

A series of 16 instances has been selected for the competition.
For generating an \x3 instance (file), you can execute for example:
\begin{command}
python BlockModeling.py -data=inst.json
python BlockModeling.py -data=[inst.json,k=3]  // if you want to change the value of k
\end{command}
where `inst.json' is a data file in JSON format.

\subsection{Bus Scheduling}

\paragraph{Description.}

From \href{https://www.csplib.org/Problems/prob022/}{CSPLib}:

\begin{quote}
Bus driver scheduling can be formulated as a set partitioning problem.
These consist of a given set of tasks (pieces of work) to cover and a large set of possible shifts, where each shift covers a subset of the tasks
and has an associated cost. We must select a subset of possible shifts that covers each piece of work once and only once: this is called a partition.
\end{quote}

\paragraph{Data.}

As an illustration of data specifying an instance of this problem, we have:
\begin{json}
{
  "nTasks": 24,
  "shifts": [
    [11, 18],
    [3, 4, 11],
    [11, 18, 19],
    [11, 12, 14, 15],
    [11, 18, 19, 20],
    [11, 12, 19, 20],
    ...   
    [21, 23],
    [0, 21]
  ]
}
\end{json}

\paragraph{Model.}

The \p3 model, in a file `BusScheduling.py', used for the \x3 competition is:

\begin{boxpy}\begin{python}
@\imp@

nTasks, shifts = data or load_json_data("t1.json")

nShifts = len(shifts)

# x[i] is 1 iff the ith shift is selected
x = VarArray(size=nShifts, dom={0, 1})

satisfy(
   # each task is covered by exactly one shift
   ExactlyOne(x[i] for i, shift in enumerate(shifts) if t in shift) for t in range(nTasks)
)

minimize(
   # minimizing the number of shifts
   Sum(x)
)
\end{python}\end{boxpy}

This model involves an array of variables and (global) constraints of type \gb{Count} and \gb{Sum}.
A series of 12 instances has been selected for the competition.
For generating an \x3 instance (file), you can execute for example:
\begin{command}
python BusScheduling.py -data=inst.json
\end{command}
where `inst.json' is a data file in JSON format.

\subsection{Button Scissors}

\paragraph{Description.}

From \href{https://github.com/lpcp-contest/lpcp-contest-2021/tree/main/problem-2}{LPCP Contest 2021}:

\begin{quote}
There is a supply of buttons attached to patches, and we have to cut them out in order to complete some very expensive suits.
Buttons in the same patch are also of mixed color, and therefore they must be properly separated after detaching.
We have to detach buttons of the same color with a cut on cardinal directions or along diagonals.
A cut must involve at least two buttons, and all buttons along the cut are detached, so they must have the same color.
Given a patch with buttons, find a sequence of cuts that detaches all buttons.
\end{quote}

\paragraph{Data.}

As an illustration of data specifying an instance of this problem, we have:
\begin{json}
{
  "nColors": 5,
  "patch": [
    [1, 2, 2, 1, 3],
    [1, 2, 4, 3, 4],
    [4, 1, 4, 5, 4],
    [3, 2, 3, 3, 4],
    [5, 5, 5, 5, 4]
  ]
}
\end{json}

\paragraph{Model.}

The \p3 model, in a file `ButtonScissors.py', used for the \x3 competition is:

\begin{boxpy}\begin{python}
@\imp@

nColors, patch = data or load_json_data("01.json")

n = len(patch)
nDiagonalTypes = 2 * n - 3  # number of possible downward diagonals 
nCodes = 2 * n + 2 * nDiagonalTypes

Attack = namedtuple("Attack", ["code", "between"])

def attack(k1, k2):
   """
   Returns a pair composed of an attack code and a list of the cells between k1 and k2
   (together with a Boolean indicating if this is the same value as in the end cells)
   """
   i1, j1, i2, j2 = k1 // n, k1 % n, k2 // n, k2 % n
   if k1 >= k2 or patch[i1][j1] != patch[i2][j2]:
      return None  # we only need to reason with k1 < k2
   v = patch[i1][j1]
   if i1 == i2:  # same row (and we know that j1 < j2)
      return Attack(i1, [(i1 * n + j, patch[i1][j] == v) for j in range(j1 + 1, j2)])
   if j1 == j2:  # same column (and we know that i1 < i2)
      return Attack(n + j1, [(i * n + j1, patch[i][j1] == v) for i in range(i1 + 1, i2)])
   if abs(i1 - i2) == abs(j1 - j2):  # same diagonal (and we know that i1 < i2)
      if j1 < j2:  # downward diagonal
         return Attack(
            2 * n + (j1 - i1 + n - 2),
            [((i1 + k) * n + j1 + k, patch[i1 + k][j1 + k] == v) for k in range(1, i2 - i1)]
         )
      else:  # upward diagonal
         return Attack(
            2 * n + nDiagonalTypes + i2 + j2 - 1,
            [((i2 - k) * n + j2 + k, patch[i2 - k][j2 + k] == v) for k in range(1, i2 - i1)]
         )
   return None

attacks = [[attack(k1, k2) for k2 in range(n * n)] for k1 in range(n * n)]
P = [(k1, k2) for k1, k2 in combinations(n * n, 2) if attacks[k1][k2]]
nMaxCuts = len(P)

# x[i][j] is the time at which the cell is cut
x = VarArray(size=n * n, dom=range(nMaxCuts))

# c[i][j] is the code of the cut for cell (i,j)
c = VarArray(size=n * n, dom=range(nCodes))

satisfy(
   # unreachable cells cannot be cut together
   [x[k1] != x[k2] for k1, k2 in combinations(n * n, 2) if (k1, k2) not in P],

   # ensuring that cuts can be performed
   [
      either(
         x[k1] != x[k2],
         x[k3] <= x[k1] if same else x[k3] < x[k1]
      ) for k1, k2 in P for k3, same in attacks[k1][k2].between
   ],

   # cells cut together (at the same time) must have the same cut code
   [
      either(
         x[k1] != x[k2],
         both(c[k1] == c[k2], c[k1] == attacks[k1][k2].code)
      ) for k1, k2 in P
   ],

   # a cut must involve at least two buttons
   [Count(within=x, value=v) != 1 for v in range(nMaxCuts)]
)

minimize(
   # minimizing the number of cuts
   Maximum(x)
)
\end{python}\end{boxpy}

This model involves two arrays of variables and (global) constraints of type \gb{Count} and \gb{Maximum}.
A series of 11 instances has been selected for the competition.
For generating an \x3 instance (file), you can execute for example:
\begin{command}
python ButtonScissors.py -data=inst.json
\end{command}
where `inst.json' is a data file in JSON format.

\subsection{Champions League}

\paragraph{Description.}

In the new organization (since 2024) of the Football Champions's league, how many points a team can get in Phase 1 (playing 8 matches) while being ranked at a given position?

\paragraph{Data.}

As an illustration of data specifying an instance of this problem, we have:
\begin{json}
{
  "schedule": [
    [["Juventus", "PSV"], ..., ["Brest", "Sturm Graz"]],
    [["Salzburg", "Brest"],  ..., ["Sturm Graz", "Club Brugge"]],
    [["Milan", "Club Brugge"],  ..., ["Young Boys", "Inter"]],
    [["PSV", "Girona"], ..., ["Bayern Munchen", "Benfica"]],
    [["S. Bratislava", "Milan"], ..., ["Bologna", "Lille"]],
    [["GNK Dinamo", "Celtic"], ..., ["Stuttgart", "Young Boys"]],
    [["Atalanta", "Sturm Graz"], ..., ["Milan", "Girona"]],
    [["Sporting CP", "Bologna"], ..., ["Brest", "Real Madrid"]]
  ]
}
\end{json}

\paragraph{Model.}

The \p3 model, in a file `ChampionsLeague.py', used for the \x3 competition is:

\begin{boxpy}\begin{python}
@\imp@

assert not variant() or variant("strict")

full_schedule, position = data or (load_json_data("phase1-2024.json"), 15)  
teams = sorted({team1 for day in full_schedule for (team1,_) in day})
assert len(teams) == 36
schedule = [[[teams.index(team1), teams.index(team2)] for (team1,team2) in day]
               for day in full_schedule]

nWeeks, nMatchesPerWeek, nTeams = len(schedule), len(schedule[0]), len(schedule[0]) * 2
W, T = range(nWeeks), range(nTeams)
assert nWeeks == 8 and nTeams == 36  # for the moment

WON, DRAWN, LOST = results = range(3)

# x[w][k] is the result of the kth match in the wth week
x = VarArray(size=[nWeeks, nMatchesPerWeek], dom=results)

# y[w][i] is the number of points won by the ith team in the wth week
y = VarArray(size=[nWeeks, nTeams], dom={0, 1, 3})

# z[i] is the number of points won by the ith team
z = VarArray(size=nTeams, dom=range(3 * nWeeks + 1))

# the target team that must be ranked at the specified position
target = Var(dom=range(nTeams))

# the number of points of the target team
z_target = Var(dom=range(3 * nWeeks + 1))

# the number of teams with a score better than the target team
better_target = Var(dom=range(nTeams))

# the number of teams with a score equal to the target team
equal_target = Var(dom=range(nTeams))

satisfy(
   # computing points won for every match
   [
      Table(
         scope=(x[w][k], y[w][i], y[w][j]),
         supports= {(WON, 3, 0), (DRAWN, 1, 1), (LOST, 0, 3)}
      ) for w in W for k, (i, j) in enumerate(schedule[w])
   ],

   # computing the number of points of each team
   [z[i] == Sum(y[:, i]) for i in T],

   # tag(redundant)
   [Sum(y[w]) in range(2 * nMatchesPerWeek, 3 * nMatchesPerWeek + 1) for w in W],

   # linking the target team with  its score
   z[target] == z_target,

   # computing the number of teams that have a score better than the target team
   better_target == Sum(z[i] > z_target for i in T),

   # computing the number of teams having a score equal to the target team (including itself)
   equal_target == Sum(z[i] == z_target for i in T),

   # nb of teams with a score better than the target team is less than the specified position
   better_target < position,

   # the position of the target team must be compatible with the specified target position
   better_target == position - 1
     if variant("strict") else
   better_target + equal_target == position
)

maximize(
   # minimizing the number of points of the target team
   z_target
)
\end{python}\end{boxpy}

This model involves three arrays of variables, four stand-alone variables and (global) constraints of type \gb{Element}, \gb{Table} and \gb{Sum}.
A series of 14 instances (7 instances for the main variant, and 7 instances for the variant ``strict'') has been selected for the competition.
For generating an \x3 instance (file), you can execute for example:
\begin{command}
  python ChampionsLeague.py -data=[phase1-2024.json,position=24]
  python ChampionsLeague.py -data=[phase1-2024.json,position=8] -variant=strict
\end{command}

\subsection{Coprime}

\paragraph{Description.}

From \cite{tabID}:

\begin{quote}
Erdős and Sárközy studied a range of problems involving coprime sets.
A pair of numbers $a$ and $b$ are coprime if there is no integer $n > 1$ which is a factor of both $a$ and $b$.
The Coprime Sets problem of size $k$ is to find the smallest $m$ such that there is a subset of $k$ distinct numbers from $\{m/2, \dots, m\}$ that are pairwise coprime.
\end{quote}

\paragraph{Data.}

An integer is required to specify a specific instance.
The values used for generating the 2025 competition instances are:
\begin{quote}
8, 10, 12, 14, 16, 17, 18, 19, 20, 21, 22, 23, 24, 25, 30, 40
\end{quote}

\paragraph{Model.}

The \p3 model, in a file `Coprime.py', used for the \x3 competition is:

\begin{boxpy}\begin{python}
@\imp@

n = data or 10

D = range(2, n * n + 1)

# x[i] is the value of the ith element of the coprime set
x = VarArray(size=n, dom=D)

satisfy(
   # setting a lower-bound
   [x[i] >= x[-1] // 2 for i in range(n - 1)],

   # tag(symmetry-breaking)
   Increasing(x, strict=True),

   # ensuring that we have coprime integers
   either(
      x[i] % d != 0,
      x[j] % d != 0
   ) for i, j in combinations(n, 2) for d in D
)

minimize(
   # minimizing the highest value of the set
   x[-1]
)
\end{python}\end{boxpy}

\begin{remark}
The model, above, is close to (can be seen as the close translation of) the one proposed in \cite{tabID}.
See Experimental Data for TabID Journal Paper.
\end{remark}

This model involves an array of variables and intensional constraints.

A series of 16 instances has been selected for the competition.
For generating an \x3 instance (file), you can execute for example:
\begin{command}
python Coprime.py -data=10
\end{command}

\subsection{Cutstock}

\paragraph{Description.}

In the cutting stock problem, we are given items with associated lengths and demands.
We are further given stock pieces of equal length and an upper bound on the number of required stock pieces for satisfying the demand.
The objective is to minimize the number of used pieces.

\paragraph{Data.}

As an illustration of data specifying an instance of this problem, we have:
\begin{json}
{
  "nPieces": 7,
  "pieceLength": 10,
  "items": [
    {"length": 7, "demand": 2},
    {"length": 5, "demand": 2},
    {"length": 3, "demand": 4}
  ]
}
\end{json}

\paragraph{Model.}

The \p3 model, in a file `Cutstock.py', used for the \x3 competition is:

\begin{boxpy}\begin{python}
@\imp@

nPieces, pieceLength, items = data or load_json_data("small.json")

lengths, demands = zip(*items)
nItems = len(data.items)

# p[i] is 1 iff the ith piece of the stock is used
p = VarArray(size=nPieces, dom={0, 1})

# r[i][j] is the number of items of type j built using stock piece i
r = VarArray(size=[nPieces, nItems], dom=range(max(demands) + 1))

satisfy(
   # not exceeding possible demands
   [r[i][j] <= demands[j] for i in range(nPieces) for j in range(nItems)],

   # each item demand must be exactly satisfied
   [Sum(r[:, j]) == demand for j, demand in enumerate(demands)],

   # each piece of the stock cannot provide more than its length
   [r[i] * lengths <= p[i] * pieceLength for i in range(nPieces)],

   # tag(symmetry-breaking)
   [
      Decreasing(p),
      LexDecreasing(r)  # to be removed for MiniCOP track
   ]
)

minimize(
   # minimizing the number of used pieces
   Sum(p)
)
\end{python}\end{boxpy}

This model involves two arrays of variables and (global) constraints of type \gb{Lex} and \gb{Sum}.
A series of 15 instances has been selected for the competition.
For generating an \x3 instance (file), you can execute for example:
\begin{command}
  python Cutstock.py -data=inst.json
\end{command}
where `inst.json' is a data file in JSON format.

\subsection{FAPP}

\paragraph{Description.}

The frequency assignment problem with polarization constraints (FAPP) is an optimization problem\footnote{This is an extended subject of the CALMA European project} that was part of the ROADEF'2001 challenge.
In this problem, there are constraints concerning frequencies and polarizations of radio links.
Progressive relaxation of these constraints is explored: the relaxation level is between 0 (no relaxation) and 10 (the maximum relaxation).  
For a complete description of the problem, the reader is invited to see \href{https://roadef.org/challenge/2001/files/fapp_roadef01_rev2_tex.pdf}{https://roadef.org/challenge/2001}.

\paragraph{Data.}

As an illustration of data specifying an instance of this problem, we have:

\begin{json}
{
  "domains": [
    [1, 2, 3, ..., 99, 100],
    [25, 26, 27, 28, 29, 30, 31, 32, 33, 34, 35, 36, 37, 38, 39, 40],
    [55, 56, 57, 63, 64, 65]
  ],
  "frequencies": [1, 1, 0, 2, 0, 0, 2, 0, 0],
  "polarizations": [-1, 0, 0, 1, 0, 0, 0, 0, 0],
  "hards": [
    [1, 2, "F", "E", 36],
    [0, 1, "F", "E", 0],
    [6, 7, "F", "I", 0],
    [2, 3, "P", "E", 0],
    [1, 6, "P", "E", 0],
    [5, 6, "P", "I", 0]
  ],
  "softs": [
    [
      0,
      2,
      [46, 44, 42, 42, 40, 40, 35, 35, 35, 30, 20],
      [39, 37, 35, 35, 33, 33, 28, 28, 28, 23, 13]
    ],
    ...,
  ]
}
\end{json}

\paragraph{Model.}

The \p3 model, in a file `FAPP.py', used for the \x3 competition is:

\begin{boxpy}\begin{python}
@\imp@

domains, frequencies, polarizations, hard_constraints, soft_constraints = data 

frequencies = [domains[f] for f in frequencies]  # we skip the indirection

n, nSofts = len(frequencies), len(soft_constraints)

def soft_table(i, j, eqr, ner, short=True):  # table for a soft constraint
   OpOverrider.disable() # to hopefully speedup things
   eq_relaxation, ne_relaxation = tuple(eqr), tuple(ner)
   T = []  # we use a list instead of a set because is quite faster to process
   cache = {}
   for f1 in frequencies[i]:
      for f2 in frequencies[j]:
         distance = absPython(f1 - f2)
         key = str(distance) + "-" + str(polarizations[i]) + "-" + str(polarizations[j])  
         if key not in cache:
            suffixes = []
            for pol in range(4):
               p1 = 0 if pol < 2 else 1
               p2 = 1 if pol in {1, 3} else 0
               if (polarizations[i], p1) in [(1, 0), (-1, 1)]
                  or (polarizations[j], p2) in [(1, 0), (-1, 1)]:
                    continue
               t = eq_relaxation if p1 == p2 else ne_relaxation  
               for kl in range(12):
                  if kl == 11 or distance >= t[kl]:  # for kl=11, we suppose t[kl] = 0
                     w1 = 0 if kl == 0 or distance >= t[kl - 1] else 1
                     w2 = 0 if kl <= 1 else next((l for l in range(kl-1) if distance >= t[l]), kl - 1)
                     suffixes.append((p1, p2, kl, w1, w2))
            cache[key] = suffixes
         elif short:
            continue
         for suffix in cache[key]:
            T.append((distance, *suffix) if short else (f1, f2, *suffix))
   OpOverrider.enable()
   return T

def domain_p(i):
   if polarizations[i] == 0:
      return {0,1}
   if polarizations[i] == 1:
      return {1}
   return {0}

def domain_d(q):
    OpOverrider.disable()
    i, j = soft_constraints[q][0], soft_constraints[q][1]
    t = {absPython(f1 - f2) for f1 in frequencies[i] for f2 in frequencies[j]}
    OpOverrider.enable()
    return t    

# f[i] is the frequency of the ith radio-link
f = VarArray(size=n, dom=lambda i: frequencies[i])

# p[i] is the polarization of the ith radio-link
p = VarArray(size=n, dom=domain_p)

# k is the relaxation level to be optimized
k = Var(dom=range(12))

# v1[q] is 1 iff the qth constraint is violated when relaxing another level
v1 = VarArray(size=nSofts, dom={0, 1})

# v2[q] is the number of times the qth cons. is violated when relaxing more than one level
v2 = VarArray(size=nSofts, dom=range(11))

# d[i][j] is the distance between the frequencies of the qth soft link
d = VarArray(size=nSofts, dom=domain_d)

satisfy(
   # imperative constraints
   [
      Match(
         (c3, c4),
         Cases={
            ("F", "E"): abs(f[i] - f[j]) == gap,
            ("P", "E"): abs(p[i] - p[j]) == gap,
            ("F", "I"): abs(f[i] - f[j]) != gap,
            ("P", "I"): abs(p[i] - p[j]) != gap
         }
      ) for (i, j, c3, c4, gap) in hard_constraints
   ],    

   # computing intermediary distances
   [d[q] == abs(f[i] - f[j]) for q, (i, j, _, _) in enumerate(soft_constraints)],
   
   # soft radio-electric compatibility constraints
   [
      Table(
         scope=(d[q], p[i], p[j], k, v1[q], v2[q]),
         supports=soft_table(i, j, eqr, ner)
      ) for q, (i, j, eqr, ner) in enumerate(soft_constraints)
   ]
)

minimize(
   k * (10 * nSofts ** 2) + Sum(v1) * (10 * nSofts) + Sum(v2)
)
\end{python}\end{boxpy}

\begin{remark}
The model, above, corresponds to the variant ``aux'' that can be found at \href{https://github.com/xcsp3team/PyCSP3-models/blob/main/realistic/FAPP/FAPP.py}{github.com\-/xcsp3team/PyCSP3-models}. 
\end{remark}

This model involves five arrays of variables, one stand-alone variable and (global) constraints of type \gb{Table}.
A series of 18 instances has been selected for the competition.
For generating an \x3 instance (file), you can execute for example:
\begin{command}
python FAPP.py -data=inst.json
\end{command}
where `inst.json' is a data file in JSON format.

\subsection{Flexible Jobshop with Scenarios}
From \href{https://www.csplib.org/Problems/prob077/}{CSPLib} (Problem 077 proposed by David Hemmi, Guido Tack and Mark Wallace):

\begin{quote}
The stochastic assignment and scheduling problem is a two-stage stochastic optimisation problem with recourse.
    A set of jobs, each composed of multiple tasks, is to be scheduled on a set of machines.
    Precedence constraints ensure that tasks, which belong to the same job are executed sequentially.
    Once the processing of a task has started, it can not be interrupted (non preemptive scheduling).
    The tasks may be restricted to a sub-set of machines. No more that one task may be executed concurrently on a machine.
    The task processing time depends on the selected machine, e.g. certain machines can finish a task faster than others.
    Furthermore, the processing times are subject to uncertainty, e.g. random variables.
    Scenarios are used to describe the uncertainty.
    A scenario describes a situation where all processing times are known and a complete schedule can be created.
    We assume that all processing times are known at the beginning of the second stage.

    The problem is composed of two stages.
    In the first stage, all the tasks have to be allocated to a machine. Once the tasks are allocated, their processing time is revealed.
    In the second stage a schedule for each machine is created, with respect to the observed processing times.
    The objective is to find a task to machine assignment minimizing the expected (average) makespan over all scenarios.
    Flexible Job Shop Scheduling is more general than Job Shop Scheduling as some tasks can be run an alternative machines.
    The goal remains to find a feasible schedule minimising the makespan.
    Each job is composed of tasks and each task must be executed by exactly one among several optional operations.
    Machines and durations are given for optional operations."
\end{quote}

\paragraph{Description.}

\paragraph{Data.}

As an illustration of data specifying an instance of this problem, we have:

\begin{json}
{
  "nMachines": 6,
  "tasks": [
    [0, 1, 2, 3, 4, 5],
    [6, 7, 8, 9, 10, 11],
    [12, 13, 14, 15, 16, 17]
  ],
  "optionalTasks": [
    [0, 1],
    [2, 3],
    [4, 5, 6],
    ...,
  ],
  "option_machines": [3, 5, 2, ..., 5],
  "scenarios": {
    "first": 0,
    "last": 4,
    "weights": [1, 1, 1, ..., 1],
    "durations": [
      [82, 54, 69, ...71],
      [67, 74, 65, ..., 60],
      ...,
      [83, 71, 58, ..., 65]
    ]
  }
}
\end{json}

\paragraph{Model.}

The \p3 model, in a file `FlexibleJobshopScen.py', used for the \x3 competition is:

\begin{boxpy}\begin{python}
@\imp@

nMachines, tasks, options, option_machines, scenarios = data 

first_scen, last_scen, weights, durations = scenarios
assert first_scen == 0 and 0 < last_scen < len(durations)
nScenarios = last_scen + 1

nJobs, nTasks, nOptions = len(tasks), len(options), len(option_machines)
J, T, O, M, S = range(nJobs), range(nTasks), range(nOptions), range(nMachines), range(nScenarios)

siblings = [next(tasks[i] for i in J if t in tasks[i]) for t in T]
minDurations = [[min(durations[s][options[t]]) for t in T] for s in S]
maxDurations = [[max(durations[s][options[t]]) for t in T] for s in S]
minStarts = [[sum(minDurations[s][k] for k in siblings[t] if k < t) for t in T] for s in S]
maxStarts = [[sum(durations[s]) - sum(minDurations[s][k] for k in siblings[t] if k >= t)
                 for t in T] for s in S]
t_max = [sum(max(durations[s][o] for o in options[t]) for t in T) for s in S]

taskForOperation = [next(t for t in T if o in options[t]) for o in O]

# x[s][t] is the starting time of the task t in the scenario s
x = VarArray(size=[S, T], dom=lambda s, t: range(minStarts[s][t], maxStarts[s][t] + 1))

# d[s][t] is the duration of the task t in the scenario s
d = VarArray(size=[S, T], dom=lambda s, t: range(minDurations[s][t], maxDurations[s][t] + 1))

# b[o] is 1 iff the optional operation o (for some task) is executed
b = VarArray(size=nOptions, dom={0, 1})

# z[s] is the duration of scenario s
z = VarArray(size=nScenarios, dom=range(max(t_max) + 1))

satisfy(
   # respecting precedence relations
   [
      x[s][t] + d[s][t] <= x[s][t + 1]
      for s in S for j in J for t in tasks[j][:-1]
   ],

   # computing durations of tasks
   [
      If(
         len(options[t]) == 1,
         Then=b[o] == 1,
         Else=If(
            b[o],
            Then=d[s][t] == durations[s][o])
      )
      for o in O for s in S if [t := taskForOperation[o]]
   ],

   # managing optional operations
   [
      [
         Sum(b[o] for o in options[t]) <= 1
         for t in T if len(options[t]) > 1
      ],
      [
         Exist(b[o] for o in options[t])
         for t in T if len(options[t]) > 1
      ],
      [
         b[min(X)] == ~b[max(X)]
         for t in T if len(X := options[t]) == 2
      ]
   ],

   # cumulative resource constraints
   [
      Cumulative(
         Task(
            origin=x[s][taskForOperation[o]],
            length=durations[s][o],
            height=b[o]
         ) for o in O if option_machines[o] == m
      ) <= 1 for m in M for s in S
   ],

   # computing the objective
   [
      x[s][tasks[j][-1]] + d[s][tasks[j][-1]] <= z[s]
      for s in S for j in J
   ]
)

minimize(
   # minimizing the weighted combination of make-spans
   Sum(weights[s] * z[s] for s in S)
)
\end{python}\end{boxpy}

\begin{remark}
The model, below, is close to (can be seen as the close translation of) the one present at CSPLib.
The original MZN model was proposed by Andreas Schutt (Copyright 2013 National ICT Australia).
\end{remark}

This model involves four arrays of variables, and (global) constraints of type \gb{Count}, \gb{Cumulative} and \gb{Sum}.

A series of 15 instances has been selected for the competition.
For generating an \x3 instance (file), you can execute for example:
\begin{command}
python FlexibleJoshopScen.py -data=inst.json
\end{command}
where `inst.json' is a data file in JSON format.

\subsection{Fortress}

\paragraph{Description.}
From \href{https://github.com/lpcp-contest/lpcp-contest-2022/tree/main/problem-1}{LPCP Contest 2022}:
\begin{quote}
A map comprises nxm cells, some of them marked with a positive integer D to denote a point-of-interest that requires D free-of-walls cells around them;
more specifically, no path of length D (Manhattan distance) originating from the point-of-interest can include walls.
All point-of-interests must be inside the perimeter of the walls, the number of walls must be minimized,
and as a second optimization criteria we prefer to minimize the amount of cells inside the perimeter of the walls.
\end{quote}

\paragraph{Data.}

As an illustration of data specifying an instance of this problem, we have:
\begin{json}
{
  "grid": [
    [0, 0, 0, 0, 0, 0, 0, 0, 0, 0, 0, 0, 0, 0, 0, 0, 0, 0, 0, 0, 0, 0, 0],
    [0, 0, 0, 0, 0, 0, 0, 0, 0, 0, 0, 0, 0, 0, 0, 0, 0, 0, 0, 0, 0, 0, 0],
    [0, 0, 0, 0, 0, 0, 0, 0, 0, 0, 0, 0, 0, 0, 0, 0, 0, 0, 0, 0, 0, 0, 0],
    [0, 0, 0, 0, 0, 0, 0, 0, 0, 0, 0, 0, 0, 0, 0, 0, 0, 0, 0, 0, 0, 0, 0],
    [0, 0, 0, 0, 0, 0, 0, 0, 0, 0, 0, 0, 0, 0, 0, 0, 0, 0, 0, 0, 0, 0, 0],
    [0, 0, 0, 0, 0, 0, 0, 0, 0, 0, 0, 0, 0, 0, 0, 0, 0, 0, 0, 0, 0, 0, 0],
    [0, 0, 0, 0, 0, 0, 0, 0, 0, 0, 0, 0, 0, 0, 0, 0, 0, 0, 0, 0, 0, 0, 0],
    [0, 0, 0, 0, 0, 0, 0, 0, 0, 0, 0, 0, 0, 0, 0, 0, 0, 0, 0, 0, 0, 0, 0],
    [0, 0, 0, 0, 0, 0, 0, 0, 3, 0, 2, 0, 3, 0, 3, 0, 0, 0, 0, 0, 0, 0, 0],
    [0, 0, 0, 0, 0, 0, 0, 0, 0, 1, 0, 0, 0, 2, 0, 0, 0, 0, 0, 0, 0, 0, 0],
    [0, 0, 0, 0, 0, 0, 0, 0, 0, 0, 0, 0, 0, 0, 0, 0, 0, 0, 0, 0, 0, 0, 0],
    [0, 0, 0, 0, 0, 0, 0, 0, 6, 0, 0, 0, 0, 3, 0, 0, 0, 0, 0, 0, 0, 0, 0],
    [0, 0, 0, 0, 0, 0, 0, 0, 0, 0, 0, 0, 0, 0, 0, 0, 0, 0, 0, 0, 0, 0, 0],
    [0, 0, 0, 0, 0, 0, 0, 0, 0, 1, 0, 0, 0, 0, 2, 0, 0, 0, 0, 0, 0, 0, 0],
    [0, 0, 0, 0, 0, 0, 0, 0, 3, 0, 0, 0, 0, 1, 0, 0, 0, 0, 0, 0, 0, 0, 0],
    [0, 0, 0, 0, 0, 0, 0, 0, 0, 0, 0, 0, 0, 0, 0, 0, 0, 0, 0, 0, 0, 0, 0],
    [0, 0, 0, 0, 0, 0, 0, 0, 0, 0, 0, 0, 0, 0, 0, 0, 0, 0, 0, 0, 0, 0, 0],
    [0, 0, 0, 0, 0, 0, 0, 0, 0, 0, 0, 0, 0, 0, 0, 0, 0, 0, 0, 0, 0, 0, 0],
    [0, 0, 0, 0, 0, 0, 0, 0, 0, 0, 0, 0, 0, 0, 0, 0, 0, 0, 0, 0, 0, 0, 0],
    [0, 0, 0, 0, 0, 0, 0, 0, 0, 0, 0, 0, 0, 0, 0, 0, 0, 0, 0, 0, 0, 0, 0],
    [0, 0, 0, 0, 0, 0, 0, 0, 0, 0, 0, 0, 0, 0, 0, 0, 0, 0, 0, 0, 0, 0, 0],
    [0, 0, 0, 0, 0, 0, 0, 0, 0, 0, 0, 0, 0, 0, 0, 0, 0, 0, 0, 0, 0, 0, 0],
    [0, 0, 0, 0, 0, 0, 0, 0, 0, 0, 0, 0, 0, 0, 0, 0, 0, 0, 0, 0, 0, 0, 0]
  ]
}
\end{json}

\paragraph{Model.}

Two models have been written for the 2025 competition. The first one, in a file `Fortress1.py', is:

\begin{boxpy}\begin{python}
@\imp@

grid = data or load_json_data("03.json")

n, m = len(grid), len(grid[0])
N, M = range(n), range(m)

P = [(i, j) for i in N for j in M if grid[i][j] != 0]

left = min(j - grid[i][j] for i, j in P) - 1
right = max(j + grid[i][j] for i, j in P) + 1
top = min(i - grid[i][j] for i, j in P) - 1
bot = max(i + grid[i][j] for i, j in P) + 1

neighborhoods = [
  [(k, q) for k in N for q in M if abs(k - i) + abs(q - j) < grid[i][j]]
  for i, j in P
]

# w[i][j] is 1 iff the cell (i,j) is a wall
w = VarArray(
   size=[n, m],
   dom=lambda i, j: {0, 1} if i in range(top, bot+1) and j in range(left, right+1) else {0}
)

# f[i][j] is 1 iff it is free (we can leave from cell at coordinates (i,j))
f = VarArray(size=[n, m], dom={0, 1})

# nW is the number of walls
nW = Var(range(n * m + 1))

# nF is the number of free cells
nF = Var(range(n * m + 1))

satisfy(

   [w[i][j] == 0 for neighborhood in neighborhoods for i, j in neighborhood],

   # setting the status of the border
   [
      Table(
         scope=(w[i][j], f[i][j]),
         supports= {(0, 1), (1, 0)}
      ) for i in N for j in M if i in (0, n - 1) or j in (0, m - 1)
   ],

   # setting the status of the point of interest
   [f[i][j] == 0 for i, j in P],

   [
      Table(
         scope=(w[i][j], f.cross(i, j)),
         supports={
           (0, 0, 0, 0, 0, 0),
           (0, 1, 1, ANY, ANY, ANY),
           (0, 1, ANY, 1, ANY, ANY),
           (0, 1, ANY, ANY, 1, ANY),
           (0, 1, ANY, ANY, ANY, 1),
           (1, 0, ANY, ANY, ANY, ANY)
         }
      ) for i in N[1:-1] for j in M[1:- 1]
   ],

   # computing thr number of walls
   nW == Sum(w),

   # computing the number of free cells
   nF + Sum(f) + nW == n * m
)

minimize(
   Sum(w) * 10000 - Sum(f)
)
\end{python}\end{boxpy}

\begin{remark}
The model, above, has not been checked to exactly correspond to the LPCP Contest statement (it was written for the 2025 XCSP3 competition).
\end{remark}

This model involves two arrays of variables, two stand-alone variables, and (global) constraints of type \gb{Table} and \gb{Sum}.
A series of 8 instances has been selected for the competition.
For generating an \x3 instance (file), you can execute for example:
\begin{command}
  python Fortress1 -data=inst.json
\end{command}

\bigskip
The second model, in a file `Fortress2.py', is:

\begin{boxpy}\begin{python}
@\imp@

grid = data or load_json_data("03.json")

n, m = len(grid), len(grid[0])
assert n <= 100 and m <= 100  # otherwise, we have to change the constant 10000 by e.g., n*m

OUT, IN, WALL = 0, 1, 10000  # 10000 is enough for a grid up to 100*100

P = [(i, j) for i in range(n) for j in range(m) if grid[i][j] != 0]
in_required = [(i, j) for i in range(n) for j in range(m) if any((k, l) for k, l in P
                   if abs(k - i) + abs(l - j) < grid[k][l])]

# table used for computing states of cells
T = {
  (OUT, OUT, ANY, ANY, ANY),
  (OUT, ANY, OUT, ANY, ANY),
  (OUT, ANY, ANY, OUT, ANY),
  (OUT, ANY, ANY, ANY, OUT),
  (IN, ne(OUT), ne(OUT), ne(OUT), ne(OUT)),
  (WALL, ANY, ANY, ANY, ANY)
}

top = min(i - grid[i][j] for i, j in P)
bot = max(i + grid[i][j] for i, j in P)
left = min(j - grid[i][j] for i, j in P)
right = max(j + grid[i][j] for i, j in P)
assert top > 0 and bot < n - 1 and left > 0 and right < m - 1  


def domain_x(i, j):
   if i < top or i > bot or j < left or j > right:
      return {OUT}  # ensuring OUT outside the possible frontier for the walls
   if (i, j) in in_required:
      return {IN}  # ensuring cells at proximity of the points of interest are inside the walls
   return {OUT, IN, WALL}

# x[i][j] is the status of the cell (i,j)
x = VarArray(size=[n, m], dom=domain_x)

satisfy(
   # computing the state of cells with respect to their (cross) neighbors
   Table(
      scope=x.cross(i, j),
      supports= T
   ) for i in range(top, bot + 1) for j in range(left, right + 1)
)

minimize(
   # minimizing the number of walls and then the number of inside cells
   Sum(x)
)
\end{python}\end{boxpy}

\begin{remark}
The model, above, has not been checked to exactly correspond to the LPCP Contest statement (it was written for the 2025 XCSP3 competition).
\end{remark}

This model involves two arrays of variables, two stand-alone variables, and (global) constraints of type \gb{Table} and \gb{Sum}.
A series of 8 instances has been selected for the competition.
For generating an \x3 instance (file), you can execute for example:
\begin{command}
  python Fortress2 -data=inst.json
\end{command}

\subsection{IHTC}

\paragraph{Description.}

  The Integrated Healthcare Timetabling Problem (IHTP), brings together three NP-hard problems and requires the following decisions:
  \begin{itemize}
  \item (i) the admission date for each patient (or admission postponement to the next scheduling period),
  \item (ii) the room for each admitted patient for the duration of their stay,
  \item (iii) the nurse for each room during each shift of the scheduling period, and (iv) the operating theater (OT) for each admitted patient.
  \end{itemize}
  See \href{ihtc2024.github.i}{ihtc2024.github.io}

\paragraph{Data.}

As an illustration of data specifying an instance of this problem, we have:

\begin{json}
{
  "nDays": 14,
  "nSkills": 3,
  "shift_types": ["early", "late", "night"],
  "age_groups": ["infant", "adult", "elderly"],
  "occupants": [
    {
      "id": "a0",
      "gender": "B",
      "age_group": "adult",
      "length_of_stay": 1,
      "workload_produced": [1, 1, 1],
      "skill_level_required": [1, 2, 0],
      "room_id": "r3"
    },
    ...
  ],
  "patients": [
    {
      "id": "p00",
      "mandatory": false,
      "gender": "A",
      "age_group": "elderly",
      "length_of_stay": 7,
      "surgery_release_day": 4,
      "surgery_due_day": 100000,
      "surgery_duration": 240,
      "surgeon_id": "s0",
      "incompatible_room_ids": ["r1"],
      "workload_produced": [2, 2, 1, ..., 1],
      "skill_level_required": [2, 1, 1, ..., 1]
    },
    ...
  ],
  "surgeons": [
    {
      "id": "s0",
      "max_surgery_time": [0, 0, 600, ..., 0]
    }
  ],
  "theaters": [
    {
      "id": "t0",
      "availability": [480, 600, 600, ..., 600]
    },
    ...
  ],
  "rooms": [
    {
      "id": "r0",
      "capacity": 3
    },
    ...
  ],
  "nurses": [
    {
      "id": "n00",
      "skill_level": 2,
      "working_shifts": [
        {"day": 0, "shift": "late", "max_load": 15},
        {"day": 2, "shift": "early", "max_load": 15},
        {"day": 3, "shift": "early", "max_load": 15},
        {"day": 4, "shift": "late", "max_load": 15},
        {"day": 5, "shift": "late", "max_load": 5},
        {"day": 6, "shift": "night", "max_load": 15},
        {"day": 8, "shift": "early", "max_load": 15},
        {"day": 9, "shift": "late", "max_load": 5},
        {"day": 10, "shift": "late", "max_load": 15},
        {"day": 11, "shift": "late", "max_load": 15},
        {"day": 12, "shift": "late", "max_load": 5}]
    },
    ...
  ],
  "weights": {
    "room_mixed_age": 5,
    "room_nurse_skill": 10,
    "continuity_of_care": 1,
    "nurse_eccessive_workload": 10,
    "open_operating_theater": 30,
    "surgeon_transfer": 10,
    "patient_delay": 10,
    "unscheduled_optional": 350
  }
}
\end{json}

\paragraph{Model.}

The \p3 model, in a file `IHTC.py', used for the \x3 competition is:

\begin{boxpy}\begin{python}
@\imp@

nDays, nSkills, shifts, ages, occupants, patients, surgeons, theaters, rooms, nurses, weights
  = data or load_json_data("i01.json")

nShifts, nPatients, nSurgeons, nTheaters, nRooms, nNurses
  = nDays * 3, len(patients), len(surgeons), len(theaters), len(rooms), len(nurses)

P, D, R = range(nPatients), range(nDays), range(nRooms)

GENDERS, A, B = ["A", "B"], 0, 1
OCCUPANTS, PATIENTS, SURGEONS, THEATERS, ROOMS, NURSES = ALL = [[obj.id for obj in t]
   for t in (occupants, patients, surgeons, theaters, rooms, nurses)]

DUMMY_DAY, DUMMY_ROOM, DUMMY_THEATER, DUMMY_NURSE = nDays, nRooms, nTheaters, nNurses
MAX = 100_000_000

max_stay = max(patient.length_of_stay for patient in patients)
shift_nurses = [[[i for i in range(nNurses) if any(asg.day == d and asg.shift == s
                   for asg in nurses[i].working_shifts)] for s in shifts] for d in D]
surgery_times = [sum(surgeon.max_surgery_time[d] for surgeon in surgeons) for d in D]
nb_min_unscheduled = nPatients - number_max_of_values_for_sum_le(sorted(patient.surgery_duration for patient in patients), sum(surgery_times))
maxll = ... # computation not shown here

# pd[i] is the patient admission day of the ith patient
pd = VarArray(size=nPatients, dom=range(nDays + 1))  # +1 for DUMMY_DAY

# pr[i] is the patient admission room of the ith patient
pr = VarArray(size=nPatients, dom=range(nRooms + 1))  # +1 for DUMMY_ROOM

# pt[i] is the patient operating theater of the ith patient
pt = VarArray(size=nPatients, dom=range(nTheaters + 1))  # +1 for DUMMY_THEATER

# pl[i] is the effective stay length of the ith patient
pl = VarArray(size=nPatients, dom=lambda i: range(patients[i].length_of_stay + 1))

# pdt[i] is the admission day combined with the operating theater of the ith patient
ptd = VarArray(size=nPatients, dom=range((nDays + 1) * (nTheaters + 1)))

# nrs[d][s][r] is the nurse for the room r at shift s of day d
nrs = VarArray(size=[nDays, 3, nRooms], dom=lambda d, s, r: shift_nurses[d][s])  


# gender[d][r] is the gender of people in the rth room on the dth day
gender = VarArray(
   size=[nDays + max_stay, nRooms + 1],
   dom=lambda i, j: {A, B} if i < nDays and j < nRooms else {-1}
)

satisfy(
   [nrs[d][s][r] == shift_nurses[d][s][0] for d in D for s in range(3) for r in R],

   # respecting possible admission days of optional patients
   [pd[i] >= patients[i].surgery_release_day for i in P if not patients[i].mandatory],

   # respecting possible admission days of mandatory patients
   [pd[i] in range(patients[i].surgery_release_day, patients[i].surgery_due_day + 1)
      for i in P if patients[i].mandatory],

   # assigning patients to compatible rooms
   [pr[i].not_among(T) for i in P
      if (T := [ROOMS.index(s) for s in patients[i].incompatible_room_ids])],

   # computing ptd
   [ptd[i] == pd[i] * (nTheaters + 1) + pt[i] for i in P],

   # taking gender of occupants into account
   [gender[d][r] == g for occupant in occupants for d in range(occupant.length_of_stay)
      if (g := GENDERS.index(occupant.gender), r := ROOMS.index(occupant.room_id))],

   # ensuring no gender mix
   [
      If(
         pd[i] + k < DUMMY_DAY,
         Then=gender[pd[i] + k, pr[i]] == GENDERS.index(patients[i].gender)
      ) for i in P for k in range(patients[i].length_of_stay)
   ],

   # not exceeding daily working time of surgeons
   [Sum(patients[i].surgery_duration * (pd[i] == d) for i in P
       if patients[i].surgeon_id == surgeon.id) <= surgeon.max_surgery_time[d]
           for surgeon in surgeons for d in D],  

   # not exceeding daily occupation time of theaters
   BinPacking(
      partition=ptd,
      sizes=[patients[i].surgery_duration for i in P],
      limits=[0 if k == nTheaters else theaters[k].availability[d] for d in D
                 for k in range(nTheaters + 1)] + [0] * nTheaters + [MAX])
   ),

   # handling postponed patients
   [
      Table(
         scope=(pd[i], pr[i], pt[i]),
         supports= {
            (DUMMY_DAY, DUMMY_ROOM, DUMMY_THEATER),
            (ne(DUMMY_DAY), ne(DUMMY_ROOM), ne(DUMMY_THEATER))
         }
      ) for i in P
   ],

   # computing effective stay lengths
   [pl[i] == min(patients[i].length_of_stay, DUMMY_DAY - pd[i]) for i in P],

   
   # respecting room capacities
   [
      Cumulative(
         tasks=[
            Task(
               origin=pd[i],
               length=(pr[i] == r) * pl[i],
               height=1
            ) for i in P
         ] + [
            Task(
               origin=0,
               length=occupant.length_of_stay,
               height=1
            ) for i, occupant in enumerate(occupants) if occupant.room_id == rooms[r].id
         ]
      ) <= rooms[r].capacity for r in R
   ],

   # tag(redundant)
   [
      BinPacking(
         partition=pd,
         sizes=1,
         limits=maxll + [MAX]
      ),

      Sum(pd[i] == DUMMY_DAY for i in P) >= nb_min_unscheduled
   ]
)

minimize(
   Sum(pd[i] == DUMMY_DAY for i in P if not patients[i].mandatory)
   * weights.unscheduled_optional 
)
\end{python}\end{boxpy}

\begin{remark}
The model proposed below (by C. Lecoutre) is an abridged version wrt the full problem.
\end{remark}

This model involves seven arrays of variables, and (global) constraints of type \gb{BinPacking}, \gb{Cumulative}, \gb{Element}, \gb{Sum} and \gb{Table}.

A series of 16 instances has been selected for the competition.
For generating an \x3 instance (file), you can execute for example:
\begin{command}
python IHTC.py -data=i01.json
\end{command}

\subsection{Low Autocorrelation}

\paragraph{Description.}

From \href{https://www.csplib.org/Problems/prob005/}{CSPLib}:
\begin{quote}
The objective is to construct a binary sequence length n that minimizes the autocorrelations between bits.
Each bit in the sequence takes the value +1 or -1.
This problem (and related ones) has many practical applications in communications and electrical engineering.
\end{quote}

\paragraph{Data.}

An integer is required to specify a specific instance: the length $n$ of the sequence.
The values used for generating the 2025 competition instances are:
\begin{quote}
  10, 20, 40, 60, 80, 100, 120, 150, 200, 300, 500, 800
\end{quote}

\paragraph{Model.}
The \p3 model, in a file `LowAutocorrelation.py', used for the competition is: 

\begin{boxpy}\begin{python}
@\imp@

n = data or 8

K = range(n - 1)  # note that we stop at n-1

# x[i] is the ith value of the sequence to be built.
x = VarArray(size=n, dom={-1, 1})

# y[k][i] is the ith product value required to compute the kth auto-correlation
y = VarArray(size=[n - 1, n - 1], dom=lambda k, i: {-1, 1} if i < n - k - 1 else None)

# c[k] is the value of the kth auto-correlation
c = VarArray(size=n - 1, dom=lambda k: range(-n + k + 1, n - k))

satisfy(
   [y[k][i] == x[i] * x[i + k + 1] for k in K for i in range(n - k - 1)],

   [Sum(y[k]) == c[k] for k in K]
)

minimize(
   # minimizing the sum of the squares of the auto-correlation
   Sum(c[k] * c[k] for k in K)
)
\end{python}\end{boxpy}

This model involves three arrays of variables, intensional constraints and constraints \gb{Sum}.
A series of 12 instances has been selected for the competition.
For generating an \x3 instance (file), you can execute for example:
\begin{command}
  python LowAutocorrelation.py -data=8 
\end{command}

\subsection{Metabolic Network}

\paragraph{Description.}

From \cite{MM_thesis}:

\begin{quote}
A metabolic network representing a set of metabolic reactions, is a directed hypergraph $(M, R, S)$ of nodes $M$, of hyperedges $R$, and of integer-valued or real-valued stoichiometry
weights on hyperedges $S$ (usually represented by a stoichiometry matrix).
Additionally, one might want to distinguish subsets of $M$: the set of internal metabolites $Int$ from the set of external metabolites $Ext$ which is generally defined as the set of nodes $m$ for which either the metabolite $m$ is
never consumed or the metabolite $m$ is never produced.
In other words, $M = Int \cup Ext$, with $Ext = \{m \in M \mid d^+(m) = 0 \land d^-(m) = 0\}$ and $Int = M \setminus Ext$,
where $d^+(m)$ and $d^-(m)$ respectively represent in degree and out degree in graph theory terms.
The field of dealing with analysis and construction of metabolic models is called metabolic modelling.
\end{quote}

Instances of this problem proposed by Maxime Mahout and François Fages (as in the Minizinc challenge 2024).

\paragraph{Data.}

As an illustration of data specifying an instance of this problem, we have:
\begin{json}
{
  "nReactions": 36,
  "stoichiometryMatrix": [
    [1, -1, 1, ..., 1],
    [0, 0, 0, ..., -1]
  ],
  "reversibleIndicators": [
    [0, 0, 0, ..., 0],
    [0, 0, 0, ..., 0],
    ..., 
    [0, 0, 0, ..., 1]
  ]
}
\end{json}

\paragraph{Model.}

The \p3 model, in a file `MetabolicNetwork.py', used for the \x3 competition is:

\begin{boxpy}\begin{python}
@\imp@

Reactions, stoichiometry_matrix, reversible_indicators = data or load_json_data("09.json")

iub = 50  # integer upper bound

# x[j] is the flux wrt the jth reaction
x = VarArray(size=nReactions, dom=range(iub + 1))

# z[j] is 1 if the jth reaction is done
z = VarArray(size=nReactions, dom={0, 1})

satisfy(
   # excluding trivial solution (only zeros)
   Sum(z) >= 1,

   # computing supports
   [z[j] == (x[j] > 0) for j in range(nReactions)],

   # handling steady-states
   [x * row == 0 for row in stoichiometry_matrix],

   # respecting reversibility
   [z * row <= 1 for row in reversible_indicators]
)

minimize(
   # minimizing reactions
   Sum(z)
)
\end{python}\end{boxpy}

This model involves two arrays of variables and (global) constraints of type \gb{Sum}.
A series of 15 instances has been proposed by Maxime Mahout and François Fages for the competition.
For generating an \x3 instance (file), you can execute for example:
\begin{command}
  python MetabolicNetwork.py -data=inst.json
\end{command}
where `inst.json' is a data file in JSON format.

\subsection{ROADEF Planning}

\paragraph{Description.}

From \cite{roadefp}:

\begin{quote}
The ROADEF conference is the largest French-speaking event aimed at bringing together researchers from various domains, including combinatorial optimization,
 operational research, constraint programming and industrial engineering.
This event is organized annually and welcomes around 600 participants.
ROADEF includes plenary sessions, tutorials in semi-plenary sessions, and multiple parallel sessions.
The conference also involves many working groups consisting of researchers collaborating on a national and potentially international level
 on specific themes covered by the conference, with each parallel session usually being organized by one or more of these working group.
The problem si to schedule ROADEF parallel sessions into available time slots while avoiding clashes among research working groups.  
\end{quote}

\paragraph{Data.}

As an illustration of data specifying an instance of this problem, we have:
\begin{json}
{
  "nSessions": 27,
  "nSlots": 11,
  "papersRange": [3, 4, 5],
  "nWorkingGroups": 17,
  "nSessionPapers": [11, 11, 8, 8, 8, 8, 10, 10, 13, 7, 4, 7, 10, 6, 5, 4, 5, 5, 5, 7, 4, 4, 6, 3, 4, 4, 5],
  "slotCapacities": [4, 5, 3, 3, 4, 3, 5, 4, 4, 3, 5],
  "sessionGroups": [ [9], [4], [9], [2], [3], [14], [10], [4], [12], [11], [4], [6], [7], [4], [1], [16], [13], [4], [17], [16], [17], [17], [9], [], [3, 8], [15, 5], [6, 4]],
  "forbidden": []
}

\end{json}

\paragraph{Model.}

The \p3 model, in a file `RoadefPlanning.py', used for the \x3 competition is:

\begin{boxpy}\begin{python}
@\imp@

papers_range, nGroups, nSessionPapers, slot_capacities, session_groups, forbidden, nRooms
   = data or (*load_json_data("2021.json"), 6)

nSessions, nSlots == len(nSessionPapers), len(slot_capacities)
I, T = range(nSessions), range(nSlots)

slot_capacities.sort()  # easier for reasoning and posting some constraints
decrement(forbidden)  # because we start indexing at 0
NO = nSlots

# pairs (i,j) of sessions shared by a positive number v of working groups
conflicts = [(i, j, v) for i, j in combinations(I, 2)
               if (v := len(set(session_groups[i]).intersection(set(session_groups[j])))) > 0]

# table used for channeling
channel_table = [(t, v) for t in T for v in range(papers_range[0], slot_capacities[t] + 1)]
                   + [(NO, 0)]

maxSlots = [next(t if (t + 1) * papers_range[0] > k else t + 1 for t in T
             if sum(slot_capacities[:t + 1]) >= k) for i, k in enumerate(nSessionPapers)]
gap = nRooms * sum(slot_capacities) - sum(nSessionPapers)
min_nb_rooms = [
  max(0, nRooms - (gap // slot_capacities[t] + (1 if gap % slot_capacities[t] != 0 else 0)))
  for t in T
]

def range_no():
   g = sum(sum(slot_capacities[: maxSlots[i]]) for i in I) - sum(nSessionPapers)
   min_nb_no = max(0, g // papers_range[-1]) + (1 if g // papers_range[-1] != 0 else 0)
   g = sum(sum(slot_capacities[-maxSlots[i]:]) for i in I) - sum(nSessionPapers)
   max_nb_no = g // papers_range[0] + (1 if g // papers_range[0] != 0 else 0)
   return range(min_nb_no, max_nb_no + 1)

def similar_sessions():
   avoidable_groups = [g for g in range(nGroups)
       if len([i for i in I if g in session_groups[i]]) == 1]
   m = [[g for g in session_groups[i] if g not in avoidable_groups] for i in I]
   return [(i, j) for i, j in combinations(I, 2)
             if nSessionPapers[i] == nSessionPapers[j] and m[i] == m[j]]

# x[i][k] is the kth slot used for session i (NO if unused slot)
x = VarArray(size=[nSessions, maxSlots], dom=range(nSlots + 1))

# y[i][k] is the number of papers presented in the kth slot used for session i (0 if unused)
y = VarArray(size=[nSessions, maxSlots], dom=[0] + papers_range)


satisfy(
   # channeling variables
   [
      Table(
         scope=(x[i][k], y[i][k]),
         supports=channel_table
      ) for i in I for k in range(maxSlots[i])
   ],

   # allocating the right number of papers for each session
   [Sum(y[i]) == nSessionPapers[i] for i in I],

   # not exceeding the possible number of parallel sessions
   Cardinality(
      within=x,
      occurrences={t: range(min_nb_rooms[t], nRooms + 1) for t in T} | {NO: range_no()}
   ),

   # forbidding some slots
   [x[i][k] != t for (i, t) in forbidden for k in range(maxSlots[i])],

   # ensuring different slots (while ordering them)
   [
      If(
         x[i][k + 1] != NO,
         Then=x[i][k] < x[i][k + 1]
      ) for i in I for k in range(maxSlots[i] - 1)
   ],

   # tag(symmetry-breaking)
   [x[i] <= x[j] for i, j in similar_sessions()]
)

   
minimize(
   Sum(
      both(
         x[i][k1] != NO,
         x[i][k1] == x[j][k2]
      ) * v
      for i, j, v in conflicts for k1 in range(maxSlots[i]) for k2 in range(maxSlots[j])
   )
)
\end{python}\end{boxpy}

This model involves two arrays of variables and (global) constraints of type \gb{Cardinality}, \gb{Lex}, \gb{Sum}, and \gb{Table}.
A series of 12 instances has been proposed by Maxime Mahout and François Fages for the competition.
For generating an \x3 instance (file), you can execute for example:
\begin{command}
  python RoadefPlanning.py -data=inst.json
\end{command}
where `inst.json' is a data file in JSON format.

\subsection{Roller Splat}

\paragraph{Description.}

Roller Splat is a ball game where you move the ball with your mouse over the course, and you try to fill up all the white squares with the colour of the ball.
It was created by Yello Games LTD.

\paragraph{Data.}

As an illustration of data specifying an instance of this problem, we have:
\begin{json}
{
  "grid": [
    [1, 1, 1, 1, 1, 1, 1, 1, 1],
    [1, 1, 1, 1, 0, 1, 1, 1, 1],
    [1, 1, 1, 1, 0, 1, 0, 0, 1],
    [1, 1, 1, 1, 0, 0, 0, 0, 1],
    [1, 0, 0, 0, 0, 0, 0, 1, 1],
    [1, 0, 0, 0, 0, 0, 0, 0, 1],
    [1, 1, 1, 1, 1, 1, 0, 0, 1],
    [1, 2, 0, 0, 0, 0, 0, 0, 1],
    [1, 1, 1, 1, 1, 1, 1, 1, 1]
  ]
}
\end{json}

\paragraph{Model.}

The \p3 model, in a file `RollerSplat.py', used for the \x3 competition is:

\begin{boxpy}\begin{python}
@\imp@

grid, horizon = data or load_json_data("04.json")

n, m = len(grid), len(grid[0])
N, M = range(n), range(m)

TOP, RGT, BOT, LFT = Directions = range(4)
EMPTY, WALL, BALL = range(3)

origin_x, origin_y = next(((i, j) for i in N for j in M if grid[i][j] == BALL))

def build_moves():
   def reachable(i, j, direction):
      tab = []
      if direction == TOP:
         while 0 <= i and grid[i][j] != WALL:
            tab.append((i, j))
            i -= 1
      elif direction == BOT:
         while i < n and grid[i][j] != WALL:
            tab.append((i, j))
           i += 1
      elif direction == LFT:
         while 0 <= j and grid[i][j] != WALL:
            tab.append((i, j))
            j -= 1
      elif direction == RGT:
         while j < m and grid[i][j] != WALL:
            tab.append((i, j))
            j += 1
      return None if len(tab) <= 1 else tab

   moves = []  # certainly, some moves could be discarded because they are not possible
   for i in range(n):
      for j in range(m):
         if t := reachable(i - 1, j, TOP):
            moves.append(t)
         if t := reachable(i + 1, j, BOT):
            moves.append(t)
         if t := reachable(i, j - 1, LFT):
            moves.append(t)
         if t := reachable(i, j + 1, RGT):
            moves.append(t)
   return moves

Moves = build_moves()
nMoves = len(Moves)

PaintingMoves = [[[k for k in range(nMoves) if (i, j) in Moves[k]] for j in M] for i in N]

T = [(k, q) for k in range(nMoves) for q in range(nMoves) if Moves[k][-1] == Moves[q][0]]
     + [(ANY, -1)]

Cells = [(i, j) for i in N for j in M if grid[i][j] != WALL]

# x[i][j] is the time of the painting move (in y) for cell at position (i,j)
x = VarArray(size=[n, m], dom=lambda i, j: range(horizon) if grid[i][j] != WALL else None)

# y[t] is the move (index) performed at time t (or -1 if finished)
y = VarArray(size=horizon + 1, dom=lambda t: range(-1, nMoves) if t < horizon else {-1})

# z is the time when the grid is entirely painted
z = Var(range(horizon))

satisfy(
   # the initial move depends on the position of the ball
   y[0] in {k for k in range(nMoves) if Moves[k][0] == (origin_x, origin_y)},

   # each cell is painted by a move that traverses it
   [y[x[i][j]] in PaintingMoves[i][j] for i, j in Cells],

   
   # each move is followed by a compatible one
   [(y[t], y[t + 1]) in T for t in range(horizon - 1)],

   # avoiding wasting time
   [
      If(
         y[t] == -1,
         Then=y[t + 1] == -1
      ) for t in range(horizon - 1)
   ],

   # computing z
   [
      y[z] != -1,
      y[z + 1] == -1
   ],

   # tag(symmetry-breaking)
   [
      If(
         x[i][j] > t,
         Then=y[t] not in PaintingMoves[i][j]
      ) for i, j in Cells for t in range(horizon - 1)
   ]
)

minimize(
   z
)
\end{python}\end{boxpy}

This model involves two arrays of variables, a stand-alone variable, and (global) constraints of type \gb{Element} and \gb{Table}.
A series of 12 instances has been generated for the competition.
For generating an \x3 instance (file), you can execute for example:
\begin{command}
  python RollerSplat.py -data=04.json
\end{command}

\subsection{Openshop Scheduling}

\paragraph{Description.}

From \href{https://en.wikipedia.org/wiki/Open-shop_scheduling}{Wikipedia}:

\begin{quote}
Open-shop scheduling is an optimization problem in computer science and operations research.
It is a variant of optimal job scheduling.
In a general job-scheduling problem, we are given n jobs J1, J2, ..., Jn of varying processing times,
which need to be scheduled on m machines with varying processing power, while trying to minimize the makespan
- the total length of the schedule (that is, when all the jobs have finished processing).
In the specific variant known as open-shop scheduling, each job consists of a set of operations O1, O2, ..., On
which need to be processed in an arbitrary order.
\end{quote}

\paragraph{Data.}

As an illustration of data specifying an instance of this problem, we have:
\begin{json}
{
  "durations": [
    [661, 6, 333],
    [168, 489, 343],
    [171, 505, 324]
  ]
}
\end{json}

\paragraph{Model.}

The \p3 model, in a file `SchedulingOS.py', used for the \x3 competition is:

\begin{boxpy}\begin{python}
@\imp@

durations = data or load_json_data("GP-os-01.json")
# durations[i][j] is the duration of operation/machine j for job i

horizon = sum(sum(t) for t in durations) + 1

n, m = len(durations), len(durations[0])
N, M = range(n), range(m)

# s[i][j] is the start time of the jth operation for the ith job
s = VarArray(size=[n, m], dom=range(horizon))

# d[i][j] is the duration of the jth operation of the ith job
d = VarArray(size=[n, m], dom=lambda i, j: durations[i])

# mc[i][j] is the machine used for the jth operation of the ith job
mc = VarArray(size=[n, m], dom=range(m))

# sd[i][k] is the start (dual) time of the kth machine when used for the ith job
sd = VarArray(size=[n, m], dom=range(horizon))

satisfy(
   # operations must be ordered on each job
   [Increasing(s[i], lengths=d[i]) for i in N],

   # each machine must be used for each job
   [AllDifferent(mc[i]) for i in N],

   [
      Table(
         scope=(mc[i][j], d[i][j]),
         supports=enumerate(durations[i])
      ) for j in M for i in N
   ],

   # tag(channeling)
   [sd[i][mc[i][j]] == s[i][j] for j in M for i in N],

   # no overlap on resources
   [
      NoOverlap(
         origins=sd[:, j],
         lengths=durations[:, j]
      ) for j in M
   ],

   # tag(redundant)
   [s[i][-1] + d[i][-1] >= sum(durations[i]) for i in N]
)

minimize(
   # minimizing the makespan
   Maximum(s[i][-1] + d[i][-1] for i in N)
)
\end{python}\end{boxpy}

This model involves four arrays of variables and (global) constraints of type \gb{AllDifferent}, \gb{Element}, \gb{Maximum}, \gb{NoOverlap}.
A series of 14 instances has been generated for the competition.
For generating an \x3 instance (file), you can execute for example:
\begin{command}
  python SchedulingOS.py -data=inst.json
\end{command}
where `inst.json' is a data file in JSON format.

\subsection{Tank Allocation}

\paragraph{Description.}

From \cite{tank12,tank18} (see also \href{https://www.csplib.org/Problems/prob051/models/}{CSPLib}):

\begin{quote}
The tank allocation problem involves the assignment of different cargoes (volumes of chemical products) to the available tanks of a vessel.
The constraints to satisfy are mainly segregation constraints:
\begin{itemize}
\item prevent chemicals from being loaded into certain types of tanks
\item prevent some pairs of cargoes to be placed next to each other
\end{itemize}
An ideal loading plan should maximize the total volume of unused tanks (i.e. free space).
\end{quote}

\paragraph{Data.}

As an illustration of data specifying an instance of this problem, we have:
\begin{json}
{
  "volumes": [1114, 979, 1068, 1267, 381, 508, 581, 557, 720, 1273, 593, 594, 793, 450, 826, 1491, 1527, 701, 673, 552],
  "conflicts": [
    [4, 14],
    [13, 17],
    [14, 17],
    [14, 19],
    [15, 17]
  ],
  "tanks": [
    {"capacity": 680, "impossibleCargos": [0, 9, 12], "neighbors": [1, 2]},
    {"capacity": 674, "impossibleCargos": [0, 9, 12], "neighbors": [0, 3]},
    {"capacity": 949, "impossibleCargos": [0, 9, 12], "neighbors": [0, 3]},
    {"capacity": 949, "impossibleCargos": [0, 9, 12], "neighbors": [1, 2]},
    {"capacity": 316, "impossibleCargos": [0, 9, 12], "neighbors": [5, 8]},
    {"capacity": 420, "impossibleCargos": [], "neighbors": [4, 6, 9]},
    {"capacity": 431, "impossibleCargos": [], "neighbors": [5, 7, 10]},
    {"capacity": 316, "impossibleCargos": [0, 9, 12], "neighbors": [6, 11]},
    {"capacity": 382, "impossibleCargos": [0, 9, 12], "neighbors": [4, 9, 12]},
    {"capacity": 451, "impossibleCargos": [0, 9, 12], "neighbors": [5, 8, 10,13]},
    {"capacity": 464, "impossibleCargos": [], "neighbors": [6, 9, 11, 14]},
    {"capacity": 382, "impossibleCargos": [0, 9, 12], "neighbors": [7, 10, 15]},
    {"capacity": 370, "impossibleCargos": [0, 9, 12], "neighbors": [8, 13]},
    {"capacity": 428, "impossibleCargos": [], "neighbors": [9, 12, 14]},
    {"capacity": 429, "impossibleCargos": [], "neighbors": [10, 13, 15]},
    {"capacity": 370, "impossibleCargos": [0, 9, 12], "neighbors": [11, 14]},
    {"capacity": 853, "impossibleCargos": [0, 9, 12], "neighbors": [17]},
    {"capacity": 991, "impossibleCargos": [], "neighbors": [16, 18]},
    {"capacity": 991, "impossibleCargos": [], "neighbors": [17, 19]},
    {"capacity": 853, "impossibleCargos": [0, 9, 12], "neighbors": [18]},
    {"capacity": 372, "impossibleCargos": [0, 9, 12], "neighbors": [21, 24]},
    {"capacity": 420, "impossibleCargos": [], "neighbors": [20, 22, 25]},
    {"capacity": 431, "impossibleCargos": [], "neighbors": [21, 23, 26]},
    {"capacity": 372, "impossibleCargos": [0, 9, 12], "neighbors": [22, 27]},
    {"capacity": 545, "impossibleCargos": [0, 9, 12], "neighbors": [20, 25]},
    {"capacity": 626, "impossibleCargos": [], "neighbors": [21, 24, 26]},
    {"capacity": 627, "impossibleCargos": [], "neighbors": [22, 25, 27]},
    {"capacity": 545, "impossibleCargos": [0, 9, 12], "neighbors": [23, 26]},
    {"capacity": 494, "impossibleCargos": [0, 9, 12], "neighbors": [29]},
    {"capacity": 589, "impossibleCargos": [], "neighbors": [28, 30]},
    {"capacity": 589, "impossibleCargos": [], "neighbors": [29, 31]},
    {"capacity": 494, "impossibleCargos": [0, 9, 12], "neighbors": [30]},
    {"capacity": 1017, "impossibleCargos": [0, 9, 12], "neighbors": [33]},
    {"capacity": 1017, "impossibleCargos": [], "neighbors": [32]}
  ]
}
\end{json}

\paragraph{Model.}

Two models have been written for the 2025 competition. The first one, in a file `TankAllocation1.py', is:

\begin{boxpy}\begin{python}
@\imp@

volumes, conflicts, tanks = data or load_json_data("chemical.json")

capacities, imp_cargos, neighbours = zip(*tanks)
sorted_capacities = sorted(capacities)

nCargos, nTanks = len(volumes), len(tanks)
C, T = range(nCargos), range(nTanks)

DUMMY_CARGO = nCargos
MAX = sum(capacities) + 1

conflicts_table = conflicts + [(v, u) for u, v in conflicts]

# x[i] is the cargo (type) of the ith tank (DUMMY_CARGO, if empty)
x = VarArray(size=nTanks, dom=range(nCargos + 1))

satisfy(
   # allocating a compatible cargo to each tank
   [x[i] not in imp_cargos[i] for i in T],

   # ensuring no adjacent tanks containing incompatible cargo
   [
      Table(
         scope=(x[i], x[j]),
         conflicts=conflicts_table
      ) for i in T for j in neighbours[i]
   ],

   # ensuring each cargo is shipped
   [
      Sum(capacities[i] * (x[i] == c) for i in T if c not in imp_cargos[i]) >= volumes[c]
      for c in C
   ]
)

maximize(
   # maximizing free space
   Sum(capacities[i] * (x[i] == DUMMY_CARGO) for i in T)
)
\end{python}\end{boxpy}

This model involves an array of variables and (global) constraints of type \gb{Sum} and \gb{Table}.
A series of 9 instances has been generated for the competition.
For generating an \x3 instance (file), you can execute for example:
\begin{command}
  python TankAllocation1.py -data=inst.json
\end{command}
where `inst.json' is a data file in JSON format.

\bigskip
The second model, in a file `TankAllocation2.py', is:

\begin{boxpy}\begin{python}
@\imp@

volumes, conflicts, tanks = data or load_json_data("chemical.json")

capacities, impossible_cargos, neighbours = zip(*tanks)
sorted_capacities = sorted(enumerate(capacities), key=lambda v: v[1])

nCargos, nTanks = len(volumes), len(tanks)
C, T = range(nCargos), range(nTanks)

conflicts = [(i, j) for i, j in conflicts if volumes[i] > 0 and volumes[j] > 0]

nMaxTanks = [number_of_values_for_sum_ge(sorted(capacities), volume) for volume in volumes]
imp_tanks = [[j for j in T if i in impossible_cargos[j]] for i in C]
DUMMY_TANK = -1

relevantCargoes = [i for i in C if volumes[i] != 0]
all_neighbors = sorted((i, j) for i in T for j in neighbours[i])  

def shipping_table(cargo):
   def rec_f(volume, tab, tmp, cursor_cap, cursor_tmp):
      for j in range(cursor_cap, nTanks):
         tmp[cursor_tmp] = sorted_capacities[j]
         if sum(tmp[k][1] for k in range(cursor_tmp + 1)) >= volume:
            tab.append(tuple(tmp[k][0] if k <= cursor_tmp else DUMMY_TANK
                               for k in range(len(tmp))))
         else:
            rec_f(volume, tab, tmp, j + 1, cursor_tmp + 1)
      return tab

   return rec_f(volumes[cargo], [], [DUMMY_TANK] * nMaxTanks[cargo], 0, 0)


# x[i][k] is the kth tank used for the ith cargo (-1, if not used)
x = VarArray(
       size=[nCargos, max(nMaxTanks)],
       dom=lambda i, k: None if volumes[i] == 0 or k >= nMaxTanks[i] else range(-1, nTanks)
    )

# y[j] is 1 if the jth tank is used
y = VarArray(size=nTanks, dom={0, 1})

satisfy(
   # using compatible tanks for each cargo
   [x[i][k] not in imp_tanks[i] for i in relevantCargoes for k in range(nMaxTanks[i])],

   # using different tanks
   AllDifferent(x, excepting=-1),

   # ensuring each cargo is shipped
   [x[i] in shipping_table(i) for i in relevantCargoes],

   # ensuring no adjacent tanks containing incompatible cargo
   [
      Table(
         scope=(x[i][k], x[j][q]),
         conflicts=all_neighbors
      ) for i, j in conflicts for k in range(nMaxTanks[i]) for q in range(nMaxTanks[j])
   ],

   # computing if tanks are used
   [y[j] == (j in flatten(x)) for j in T]
)

maximize(
   # maximizing free space
   Sum(capacities[j] * (y[j] == 0) for j in T)
)
\end{python}\end{boxpy}

This model involves two arrays of variables and (global) constraints of type \gb{AllDifferent}, \gb{Count}, \gb{Sum}, and \gb{Table}.
A series of 9 instances has been generated for the competition.
For generating an \x3 instance (file), you can execute for example:
\begin{command}
  python TankAllocation2.py -data=inst.json
\end{command}
where `inst.json' is a data file in JSON format

\chapter{Solvers}

In this chapter, we introduce the solvers and teams having participated to the \x3 Competition 2025.

\begin{itemize}
\item ACE (Christophe Lecoutre)
\item BTD, miniBTD {\small(M. Sami Cherif, Djamal Habet, Philippe J\'egou, H\'el\`ene Kanso, Cyril Terrioux)}
\item Choco (Charles Prud'homme)
\item CoSoCo (Gilles Audemard)
\item CPMpy {\small (Tias Guns, Thomas Sergeys, Ignace Bleukx, Dimos Tsouros, Hendrik Bierlee)} \\
  cpmpy\_ortools, cpmpy\_cpo, cpmpy\_mzn\_chuffed, cpmpy\_mzn\_gecode, \\
  cpmpy\_z3, cpmpy\_exact,
  cpmpy\_gurobi
\item Exchequer (Martin Mariusz Lester)
\item Fun-sCOP (Takehide Soh, Daniel Le Berre, Hidetomo Nabeshima, Mutsunori Banbara, Naoyuki Tamura)
\item Nacre (Ga\"el Glorian)
\item Picat (Neng-Fa Zhou)
\item PyCSP3-ortools (Nicolas Szczepanski)
\item RBO, miniRBO (Mohamed Sami Cherif, Djamal Habet, Cyril Terrioux)
\item Sat4j-CSP-PB (extension of Sat4j by Thibault Falque and Romain Wallon)
\item toulbar2 (David Allouche, Simon de Givry  et al.) 
\end{itemize}

\addcontentsline{toc}{section}{\numberline{}ACE}
\includepdf[pages=-,pagecommand={\thispagestyle{plain}}]{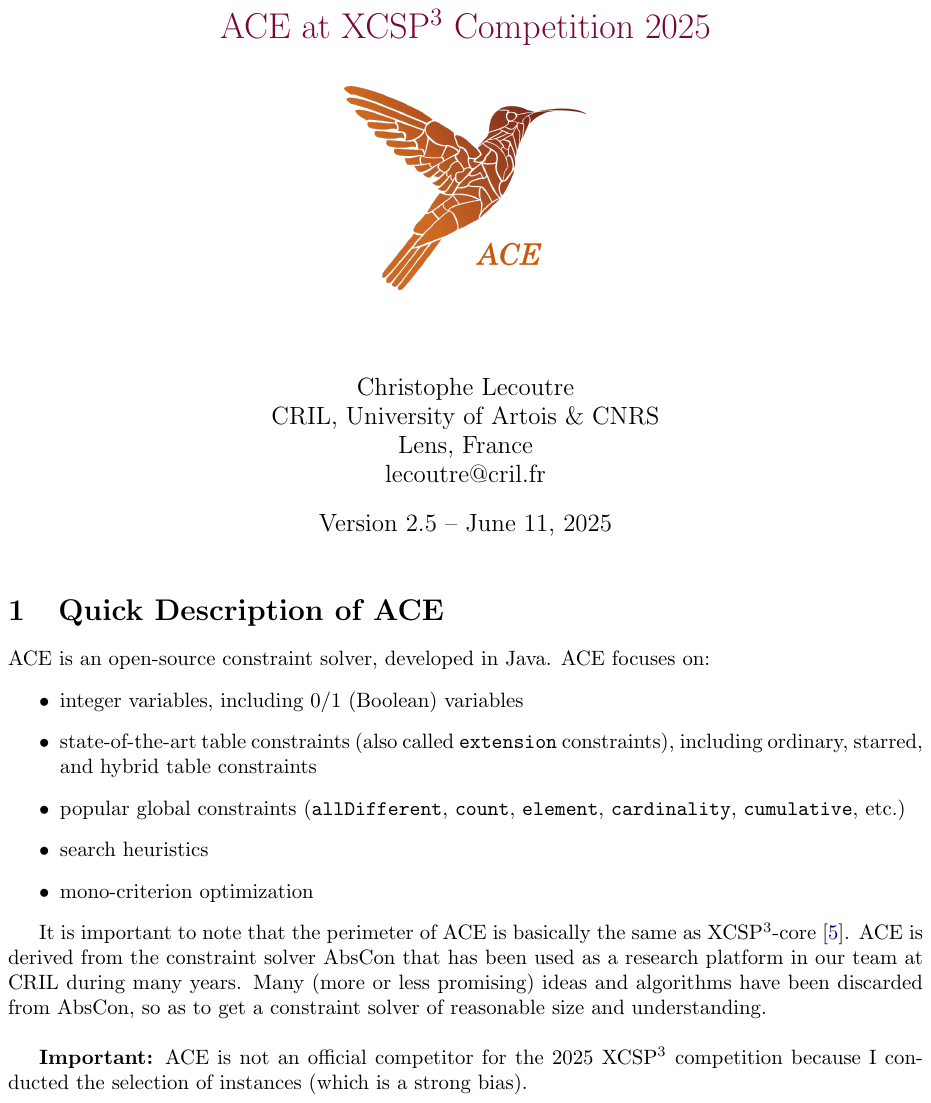}
\addcontentsline{toc}{section}{\numberline{}BTD, miniBTD}
\includepdf[pages=-,pagecommand={\thispagestyle{plain}}]{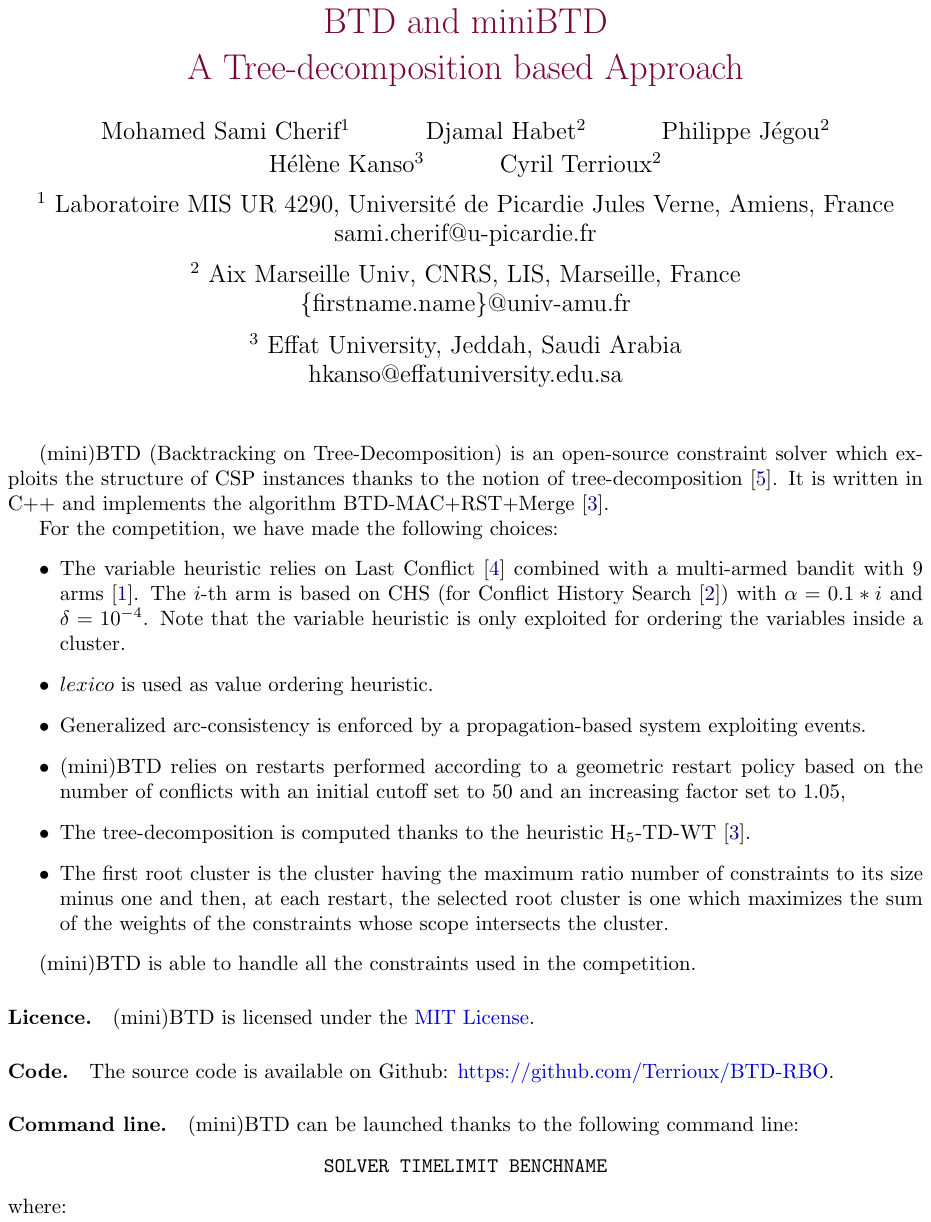}
\addcontentsline{toc}{section}{\numberline{}Choco}
\includepdf[pages=-,pagecommand={\thispagestyle{plain}}]{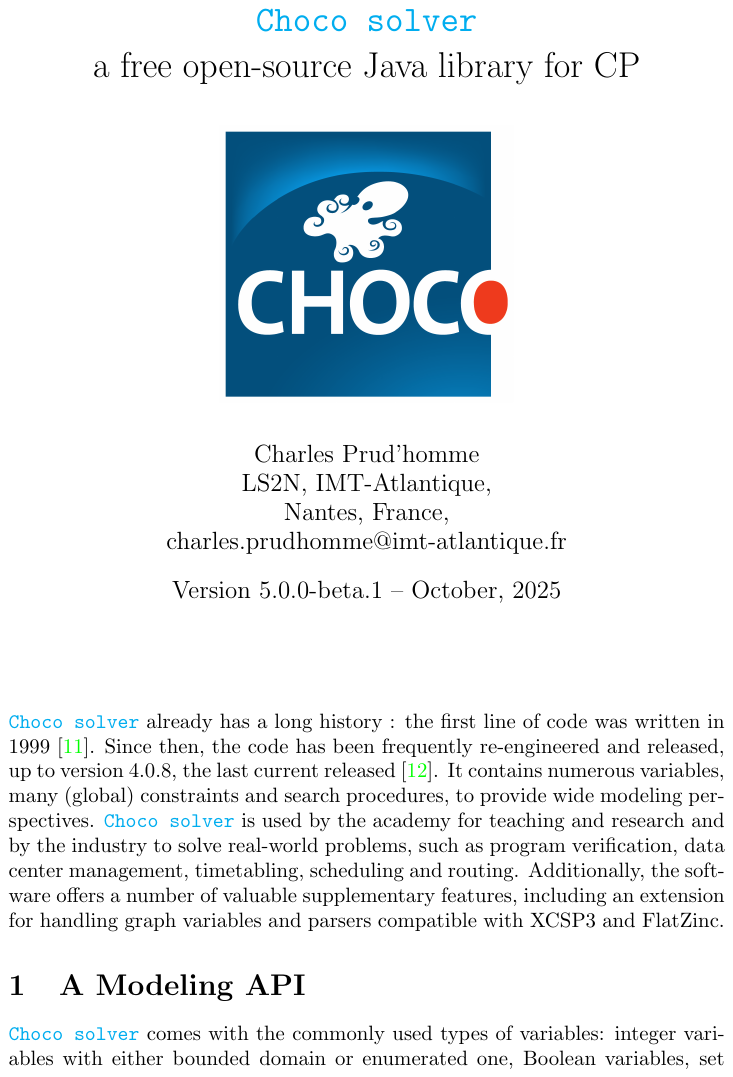}
\addcontentsline{toc}{section}{\numberline{}CoSoCo}
\includepdf[pages=-,pagecommand={\thispagestyle{plain}}]{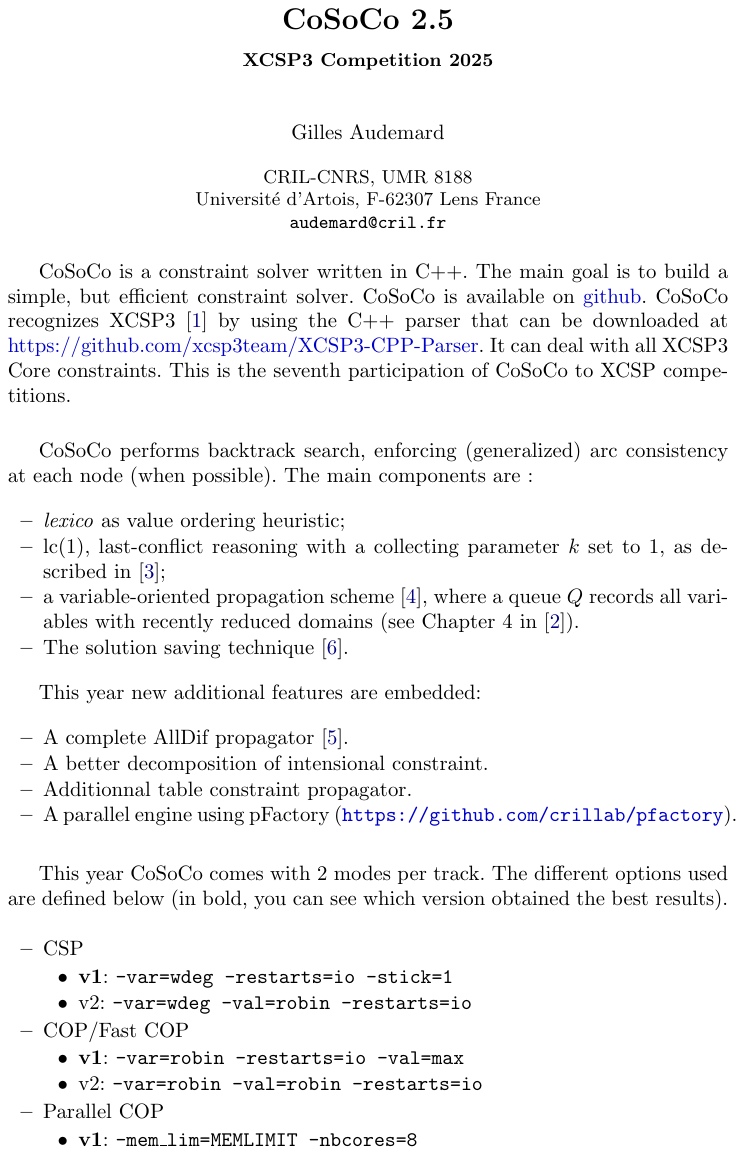}
\addcontentsline{toc}{section}{\numberline{}CPMpy-*}
\includepdf[pages=-,pagecommand={\thispagestyle{plain}}]{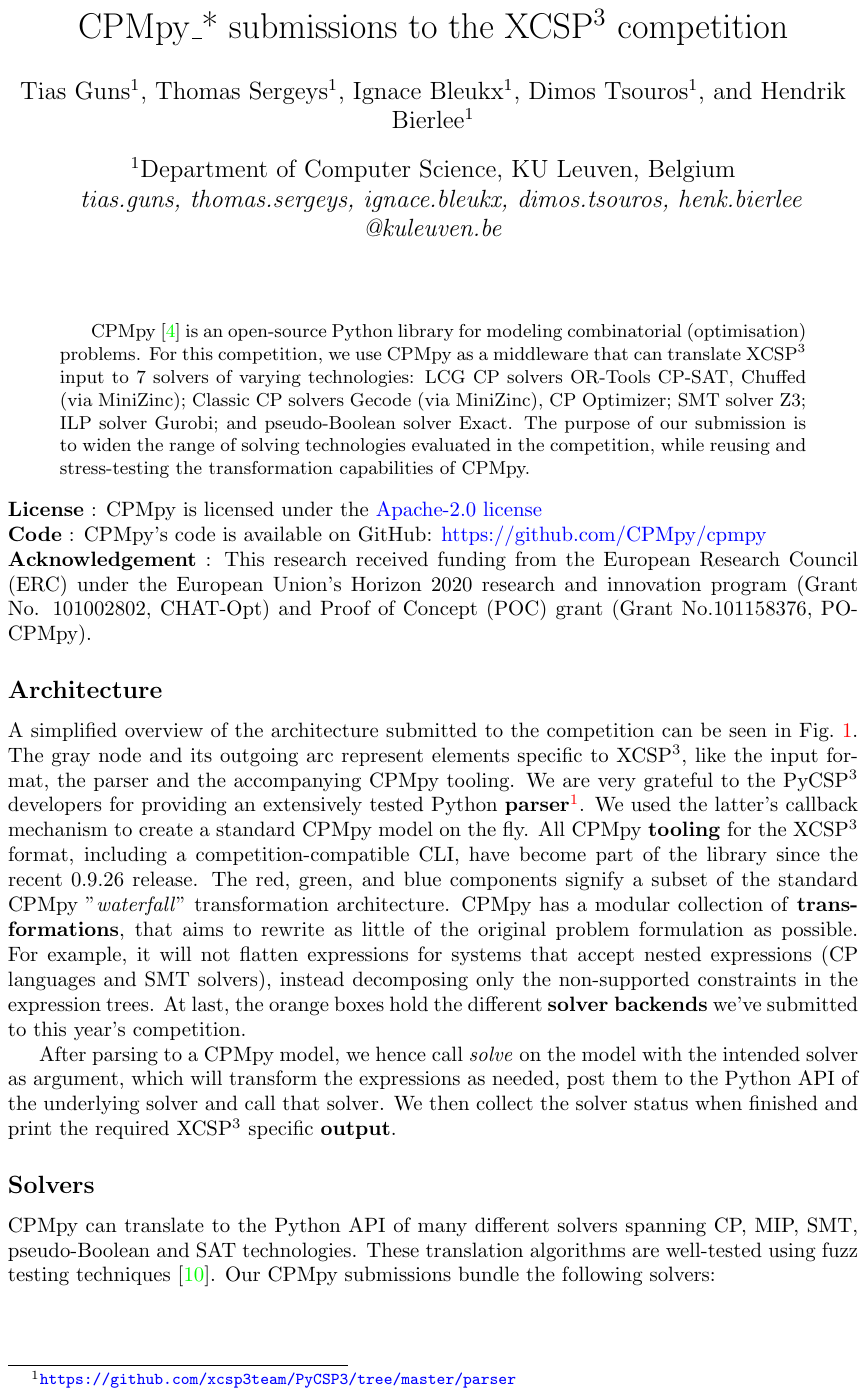}
\addcontentsline{toc}{section}{\numberline{}Exchequer}
\addcontentsline{toc}{section}{\numberline{}Fun-sCOP}
\includepdf[pages=-,pagecommand={\thispagestyle{plain}}]{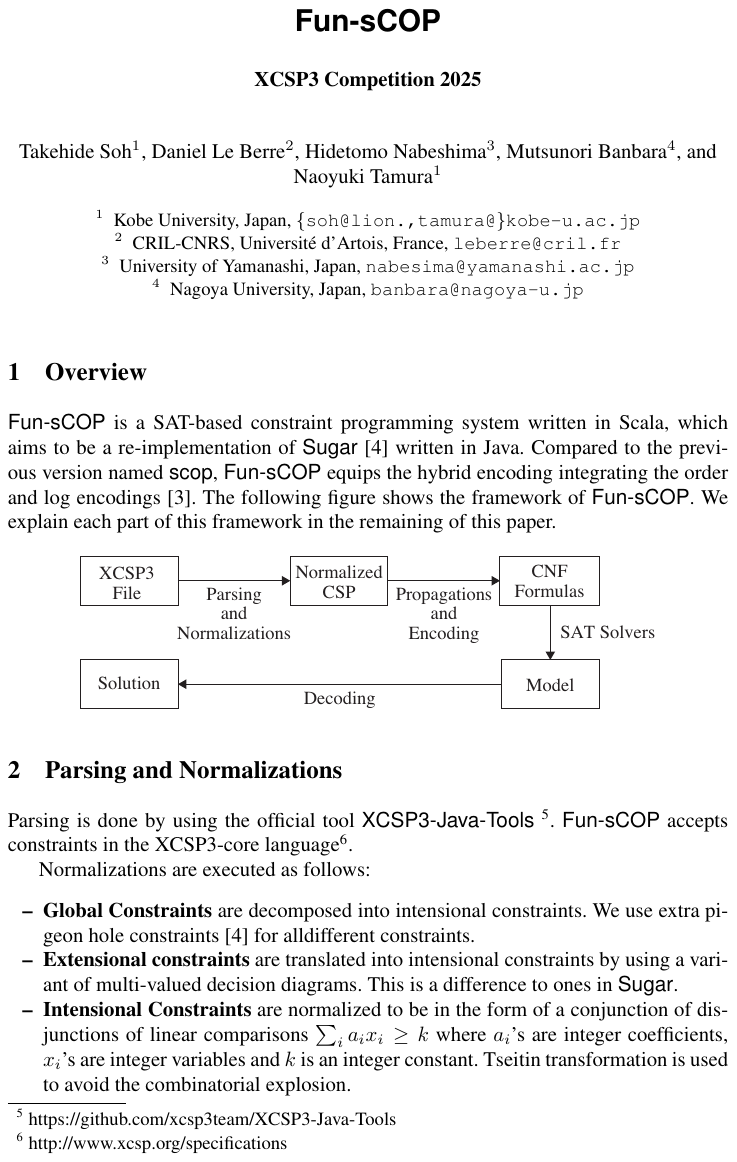}
\addcontentsline{toc}{section}{\numberline{}Nacre}
\includepdf[pages=-,pagecommand={\thispagestyle{plain}}]{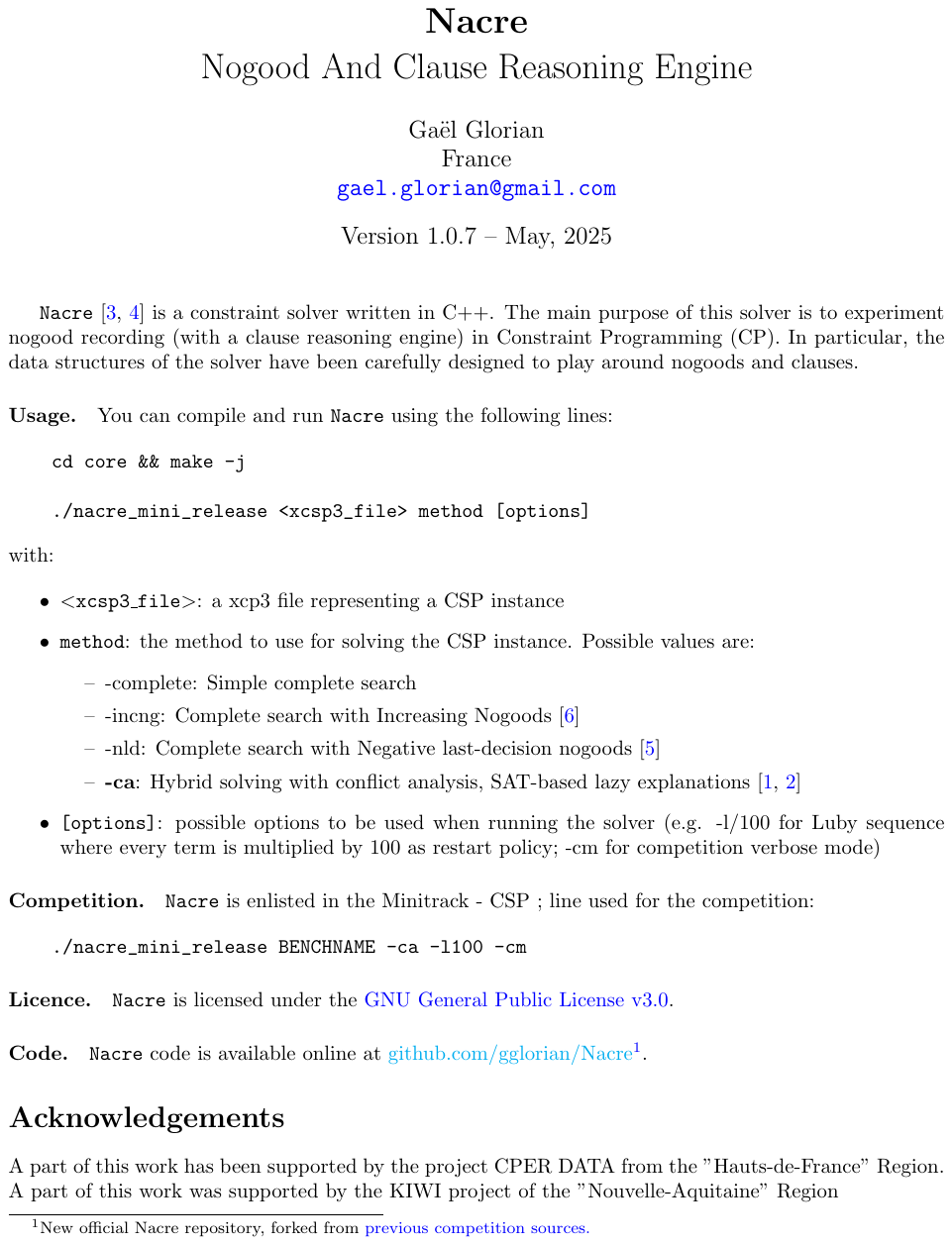}
\addcontentsline{toc}{section}{\numberline{}Picat}
\includepdf[pages=-,pagecommand={\thispagestyle{plain}}]{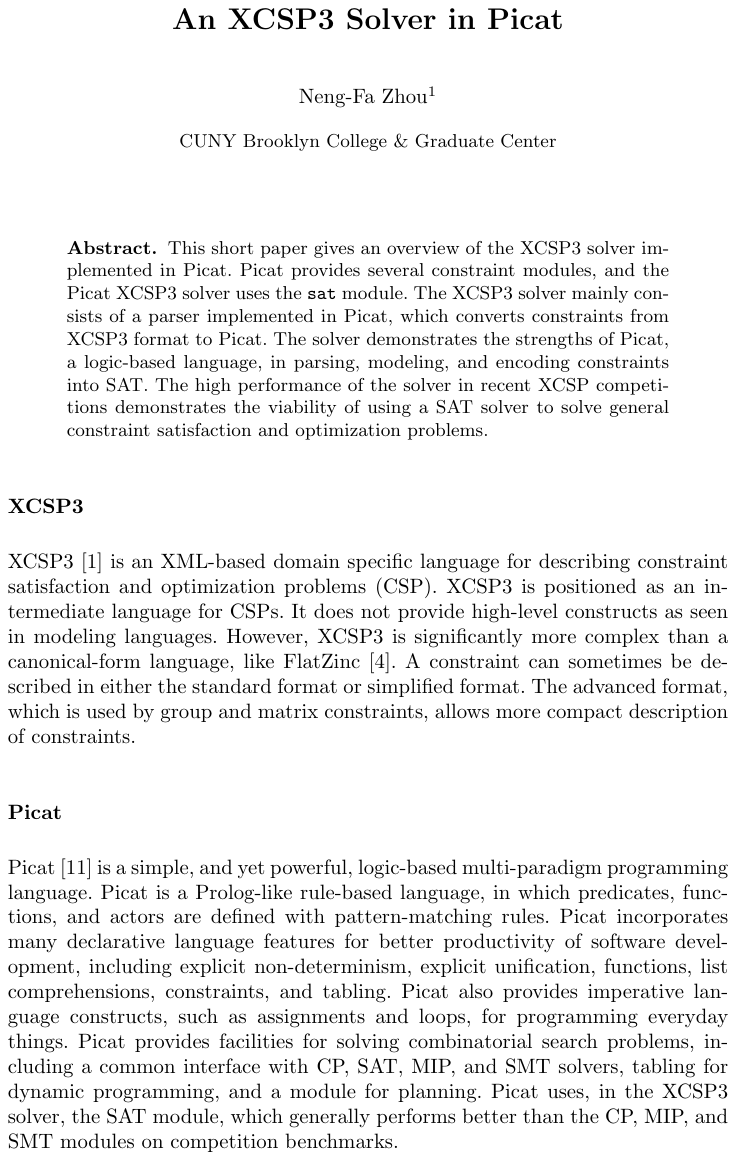}
\addcontentsline{toc}{section}{\numberline{}PyCSP3-ortools}
\includepdf[pages=-,pagecommand={\thispagestyle{plain}}]{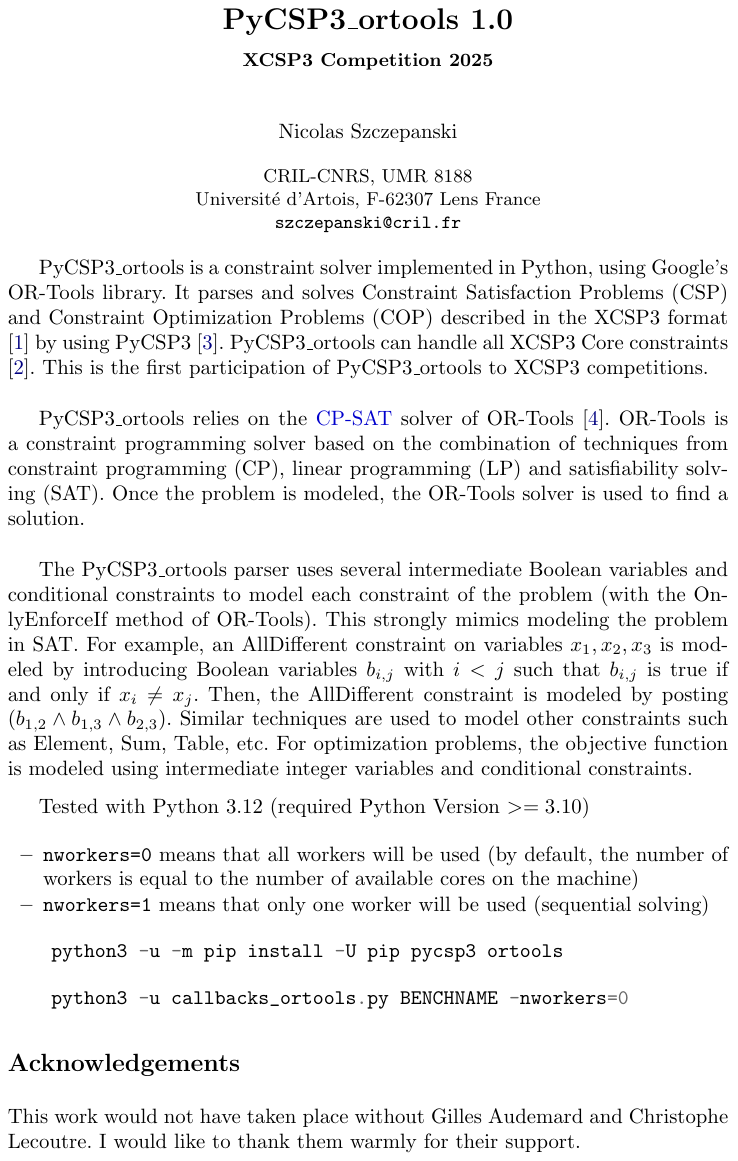}
\addcontentsline{toc}{section}{\numberline{}RBO, miniRBO}
\includepdf[pages=-,pagecommand={\thispagestyle{plain}}]{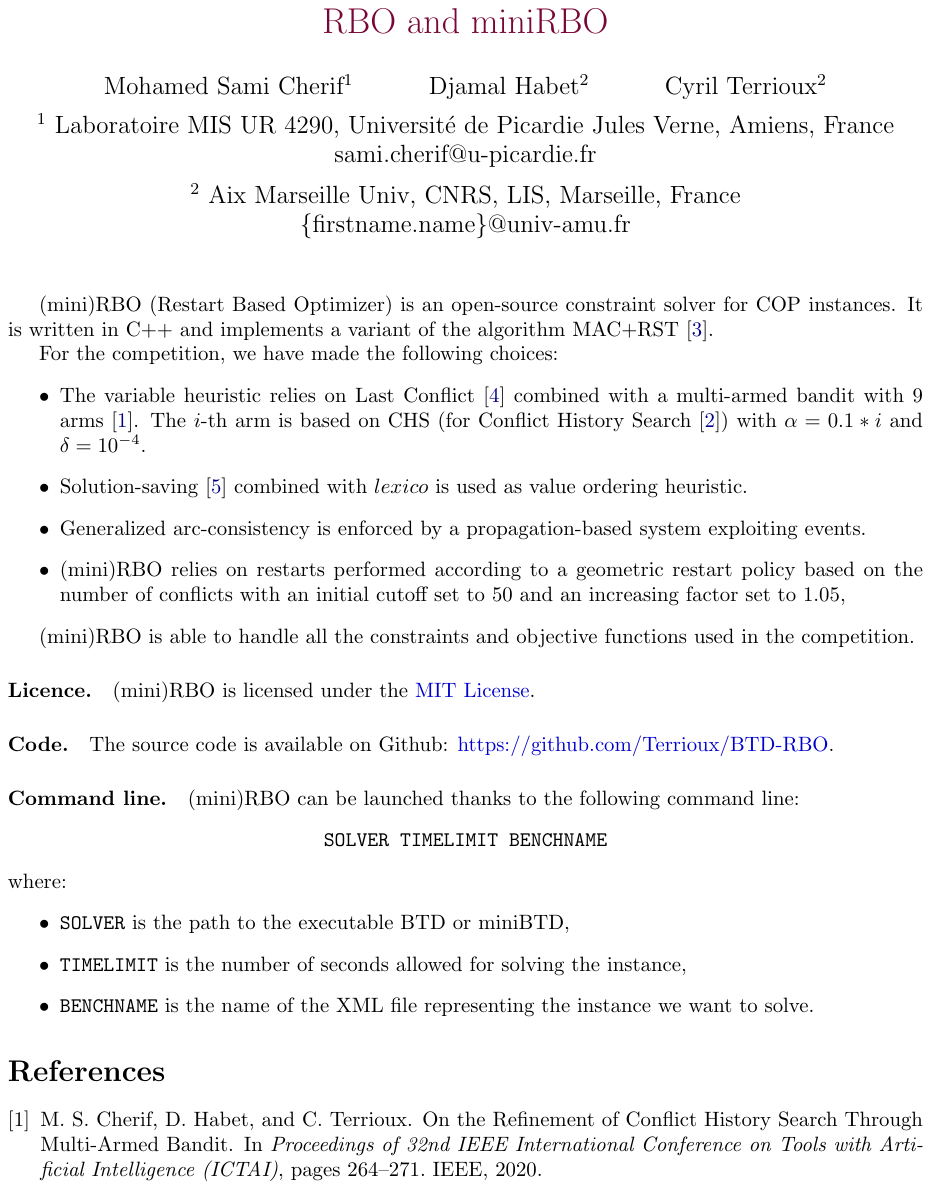}
\addcontentsline{toc}{section}{\numberline{}Sat4j-CSP-PB}
\includepdf[pages=-,pagecommand={\thispagestyle{plain}}]{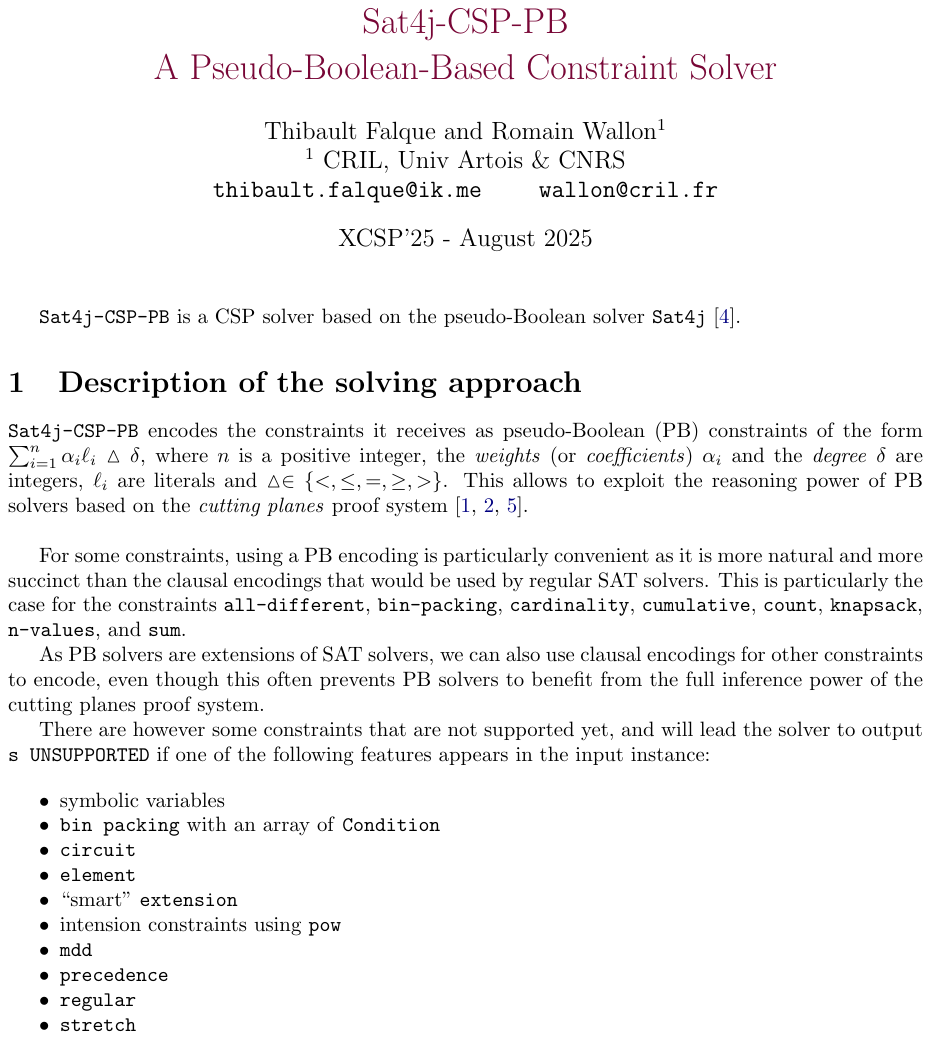}
\addcontentsline{toc}{section}{\numberline{}toulbar2}
\includepdf[pages=-,pagecommand={\thispagestyle{plain}}]{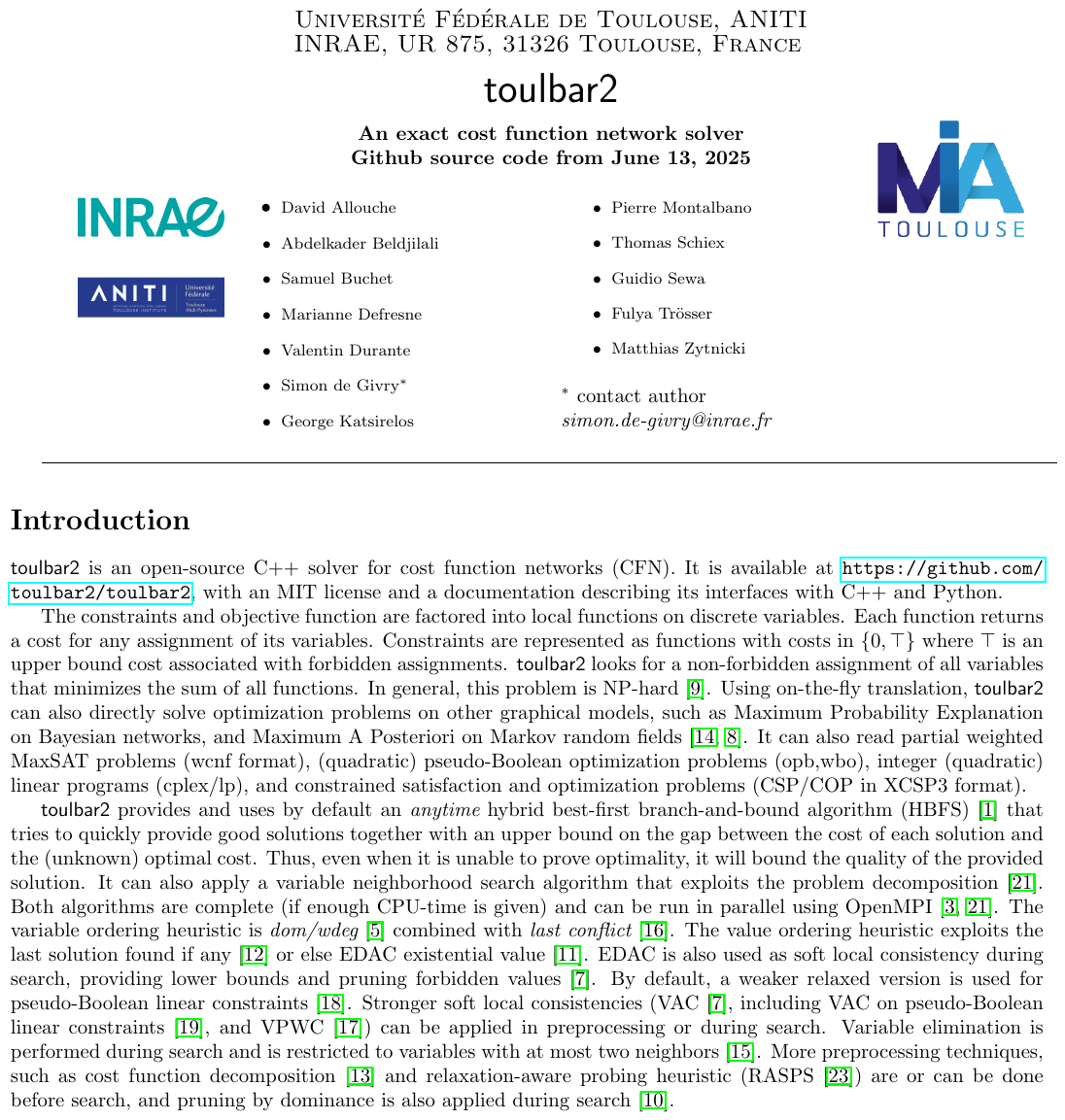}


\chapter{Results}

In this chapter, rankings for the various tracks of the \x3 Competition 2025 are given.
Importantly, remember that you can find all detailed results, including all traces of solvers at \href{https://www.cril.univ-artois.fr/XCSP25/}{https://www.cril.univ-artois.fr/XCSP25/}.

\section{Context}

\bigskip
Remember that the tracks of the competition are given by Table \ref{tab:anysolver} and Table \ref{tab:minisolver}.

\begin{table}[h!]
\begin{center}
\begin{tabular}{cccc} 
\toprule
\textcolor{dred}{\bf Problem} &  \textcolor{dred}{\bf Goal} &  \textcolor{dred}{\bf Exploration} &  \textcolor{dred}{\bf Timeout} \\
\midrule
CSP  & one solution & sequential & 30 minutes \\
COP  & best solution & sequential & 30 minutes \\
Fast COP  & best solution & sequential & 3 minutes \\
// COP  & best solution & parallel & 30 minutes \\
\bottomrule
\end{tabular}
\end{center}
\caption{Standard Tracks. \label{tab:anysolver}}
\end{table}

\begin{table}[h!]
\begin{center}
\begin{tabular}{cccc} 
\toprule
\textcolor{dred}{\bf Problem} &  \textcolor{dred}{\bf Goal} &  \textcolor{dred}{\bf Exploration} &  \textcolor{dred}{\bf Timeout} \\
\midrule
Mini CSP  & one solution & sequential & 30 minutes \\
Mini COP  & best solution & sequential & 30 minutes \\
\bottomrule
\end{tabular}
\end{center}
\caption{Mini-Solver Tracks. \label{tab:minisolver}}
\end{table}

\noindent Also, note that:

\begin{itemize}
\item The cluster was provided by CRIL and is composed of nodes with 4 quad-core Intel(R) Xeon(R) Gold 6248 CPU @ 2.50GHz, each equipped with 192GiB RAM.
\item Each solver was allocated a CPU and 64 GiB of RAM, independently from the tracks.
\item Timeouts were set accordingly to the tracks through the tool \texttt{runsolver}:
 \begin{itemize}
    \item sequential solvers in the fast COP track were allocated 3 minutes of CPU time and 4.5 minutes of Wall Clock time,
    \item other sequential solvers were allocated 30 minutes of CPU time and 45 min of Wall Clock time,
    \item parallel solvers were allocated 4 CPU and 30 minutes of Wall Clock Time.
 \end{itemize}
 \item The selection of instances for the Standard tracks was composed of 200 CSP instances and 250~COP instances.
\item The selection of instances for the Mini-solver tracks was composed of 150 CSP instances and 150~COP instances.
\end{itemize}

\paragraph{About Scoring.}
The number of points won by a solver $S$ is decided as follows:
\begin{itemize}
\item for CSP, this is the number of times $S$ is able to solve an instance, i.e., to decide the satisfiability of an instance (either exhibiting a solution, or indicating that the instance is unsatisfiable)
\item for COP, this is, roughly speaking, the number of times $S$ gives the best known result, compared to its competitors.
More specifically, for each instance $I$:
\begin{itemize}
\item if $I$ is unsatisfiable, 1 point is won by $S$ if $S$ indicates that the instance $I$ is unsatisfiable, 0 otherwise,
\item if $S$ provides a solution whose bound is less good than another one (found by another competiting solver), 0 point is won by $S$,
\item if $S$ provides an optimal solution, while indicating that it is indeed the optimality, 1 point is won by $S$,
\item if $S$ provides (a solution with) the best found bound among all competitors, this being possibly shared by some other solver(s), while indicating no information about optimality: 1 point is won by $S$ if no other solver proved that this bound was optimal, $0.5$ otherwise.
\end{itemize}
\end{itemize}


\paragraph{Off-competition Solvers.} Some solvers were run while not officially entering the competition: we call them {\em off-competition} solvers.
\ace is one of them because its author (C. Lecoutre) conducted the selection of instances, which is a very strong bias.
Also, when two or more variants (by the same competiting team) of a same solver compete in a same track, only the best one is ranked (and the other ones considered as being off-competition).
To determine which solver variant is the best for a team, we compute the ranking with only the variants of the team competiting between them. 
This is why, for example, in some cases, solvers submitted by the CPMpy team were discarded from the ranking. 

\section{Rankings}

Recall that, concerning ranking, two new rules are used when necessary:
\begin{itemize}
\item In case a team submits the same solver to both the main track and the mini-track for the same problem (CSP or COP), 
the solver will be ranked in the mini-track only if the solver is not one of the three best solvers in the main track.
\item In case several teams submit variations of the same solver to the same track, only
the team who developed the solver and the best other team with that solver will
be ranked (possibly, a second best other team, if the jury thinks that it is relevant)
\end{itemize}

\bigskip
\noindent The algorithm used in practice for establishing the ranking is:
\begin{enumerate}
\item first, off-competition solvers are discarded (this is the case for ACE in 2025)
\item second, only the best variation of the same solver (submitted by the same team) is kept when computing scores (ranking) ; to identify the best variation of a solver, we compute scores while discarding all other solvers (only variataiosn of the solver participate). 
\item third, in mini-tracks, (any variations of) solvers that (whose variations) are ranked 1st, 2nd or 3rd in the corresponding main track are discarded 
\end{enumerate}

\bigskip
Here are the rankings\footnote{The images of medals come from \href{https://www.flaticon.com/free-icons/2nd-place}{Flaticon -- icons created by Md Tanvirul Haque.}} 
     for the 6 tracks.
As PyCSP3\_ortools and  CPMpy\_ortools have obtained close results in some tracks (corresponding to the same ranking), we decided to put them together (ex-aequo in term of ranking) for these tracks. 

\bigskip 

\begin{minipage}{0.4\textwidth}
  \begin{center}
    \hfill
  \begin{tabular}{cp{3.2cm}}
    \multicolumn{2}{l}{{\bf {\Large CSP}}} \\
     \midrule
     \vspace{-0.2cm} &  \\
     \includegraphics[scale=0.075]{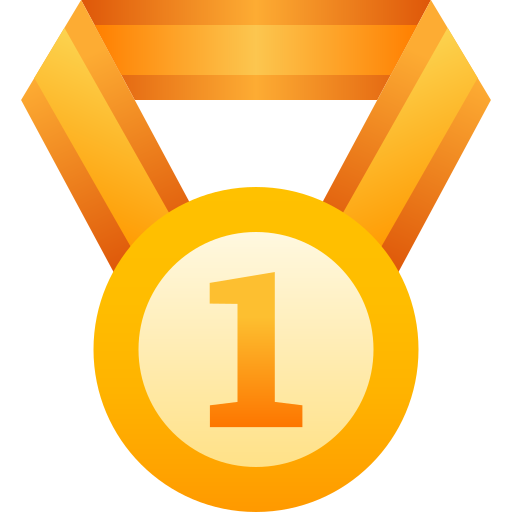} &  \vspace{-0.6cm} {\large Picat} \\
      \midrule
      & \\
      \includegraphics[scale=0.075]{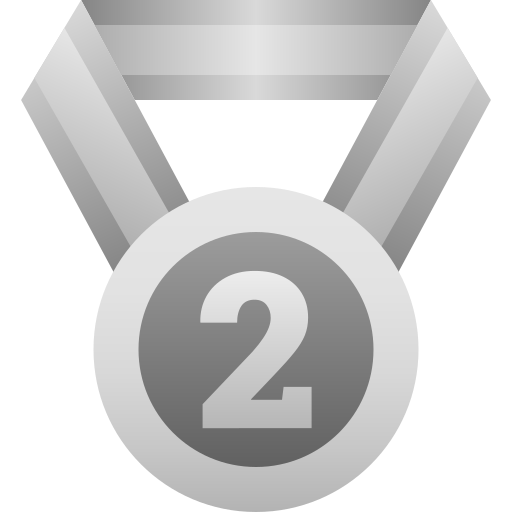} &  \vspace{-0.9cm} {\large PyCSP3\_ortools CPMpy\_ortools} \\
       \midrule
      & \\
       \includegraphics[scale=0.075]{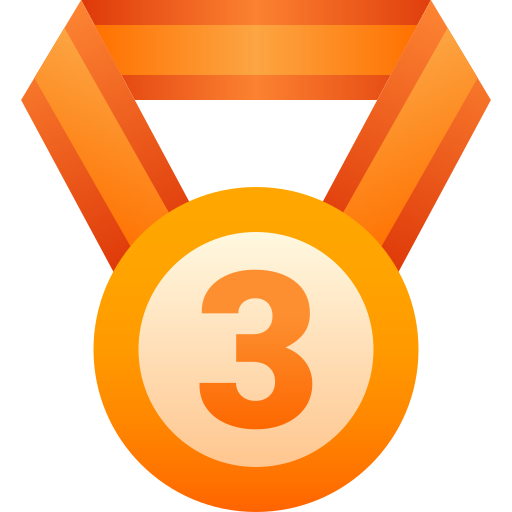} &  \vspace{-0.6cm} {\large Choco} \\
        \midrule
  \end{tabular}
\end{center}
\end{minipage} 
\hfill
\begin{minipage}{0.4\textwidth}
\begin{center}
  \begin{tabular}{cp{3.2cm}}
    \multicolumn{2}{l}{{\bf {\Large COP}}} \\
     \midrule
    \vspace{-0.2cm} &  \\
    \includegraphics[scale=0.075]{gold.png} &  \vspace{-0.6cm}  {\large CPMpy\_ortools} \\
     \midrule
      & \\
     \includegraphics[scale=0.075]{silver.png} &  \vspace{-0.6cm} {\large CoSoCo} \\
      \midrule
      & \\
      \includegraphics[scale=0.075]{bronze.png} &  \vspace{-0.6cm} {\large Picat} \\
       \midrule
  \end{tabular}
\end{center}
\end{minipage}
\hfill

\bigskip\bigskip 

\begin{minipage}{0.4\textwidth}  
  \begin{center}
    \hfill
  \begin{tabular}{cp{3.2cm}}
    \multicolumn{2}{l}{{\bf {\Large Fast COP}}} \\
     \midrule
    \vspace{-0.2cm} &  \\
    \includegraphics[scale=0.075]{gold.png} &  \vspace{-0.6cm} {\large CPMpy\_ortools} \\
     \midrule
       & \\
     \includegraphics[scale=0.075]{silver.png} &  \vspace{-0.6cm} {\large CoSoCo} \\
      \midrule
      & \\
      \includegraphics[scale=0.075]{bronze.png} &  \vspace{-0.6cm} {\large Choco} \\
       \midrule
  \end{tabular}
\end{center}
\end{minipage}
\hfill 
\begin{minipage}{0.4\textwidth}
\begin{center}
  \begin{tabular}{cp{3.2cm}}
    \multicolumn{2}{l}{{\bf {\Large Parallel COP}}} \\
     \midrule
    \vspace{-0.2cm}   &  \\
    \includegraphics[scale=0.075]{gold.png} &  \vspace{-0.9cm} {\large CPMpy\_ortools PyCSP3\_ortools} \\
     \midrule
      & \\
     \includegraphics[scale=0.075]{silver.png} &  \vspace{-0.6cm} {\large Choco} \\
      \midrule
      & \\
      \includegraphics[scale=0.075]{bronze.png} &  \vspace{-0.6cm} {\large CoSoCo}  \\
       \midrule
  \end{tabular}
\end{center}
\end{minipage}
\hfill

\bigskip\bigskip 

\begin{minipage}{0.4\textwidth}  
  \begin{center}
    \hfill
  \begin{tabular}{cp{3.2cm}}
    \multicolumn{2}{l}{{\bf {\Large Mini CSP}}} \\
    \midrule
    \vspace{-0.2cm} &   \\
    \includegraphics[scale=0.075]{gold.png} &  \vspace{-0.6cm} {\large miniBTD} \\
    \midrule
      & \\
    \includegraphics[scale=0.075]{silver.png} &  \vspace{-0.6cm} {\large Nacre} \\
    \midrule
      & \\
    \includegraphics[scale=0.075]{bronze.png} &  \vspace{-0.6cm} {\large Exchequer} \\
    \midrule
  \end{tabular}
\end{center}
\end{minipage}
\hfill 
\begin{minipage}{0.4\textwidth}
\begin{center}
  \begin{tabular}{cp{3.2cm}}
     \multicolumn{2}{l}{{\bf {\Large Mini COP}}} \\
     \midrule
    \vspace{-0.2cm} &  \\
    \includegraphics[scale=0.075]{gold.png} &  \vspace{-0.6cm} {\large Toulbar2} \\
      \midrule
      & \\
      \includegraphics[scale=0.075]{silver.png} &  \vspace{-0.6cm} {\large Sat4j} \\
        \midrule
      & \\
        \includegraphics[scale=0.075]{bronze.png} &  \vspace{-0.6cm} {\large miniRBO} \\
          \midrule
  \end{tabular}
\end{center}
\end{minipage}
\hfill


\end{document}